\PassOptionsToPackage{dvipsnames,svgnames,x11names}{xcolor}
\documentclass[12pt]{article}

\usepackage{amsmath,amsthm,amssymb}
\usepackage{mathrsfs}
\usepackage{bm}
\allowdisplaybreaks 

\usepackage{graphicx}
\usepackage[ruled,vlined]{algorithm2e}
\usepackage{float}
\usepackage{lscape}
\usepackage{booktabs}
\usepackage{multirow}
\usepackage{threeparttable}
\usepackage{caption}
\usepackage{subcaption}
\usepackage{enumerate}
\usepackage{verbatim}
\usepackage{xcolor}

\usepackage[round]{natbib}
\usepackage{hyperref}
\hypersetup{
  colorlinks=true,
  linkcolor=blue,
  citecolor=blue,
  urlcolor=blue,
  anchorcolor=blue}

\newtheorem{theorem}{Theorem}[section]
\newtheorem{lemma}{Lemma}[section]
\newtheorem{lemma*}{Lemma}
\newtheorem{remark}{Remark}[section]
\newtheorem{remark*}{Remark}

\newtheorem{assumption}{Assumption}[section]

\DeclareMathOperator*{\argmin}{argmin}

\def\m0{\mathbf{0}}

\def \mI {\mathbf{I}}

\def \mrd {\mathrm{d}}
\def \mrN {\mathrm{NN}}

\def \mcL {\mathcal{L}}
\def \mcN {\mathcal{N}}

\def\wh{\widehat}
\def\wt{\widetilde}

\newcommand{\anon}{1}

\date{}

\newcommand{\paperTitle}{Conditional Diffusion for Nonparametric Instrumental Variable Quantile Regression}

\begin{document}

\def\spacingset#1{\renewcommand{\baselinestretch}{#1}\small\normalsize}
\spacingset{1}


\if1\anon
{
  \title{\bf \paperTitle}
\author{
Xingdong Feng 
\thanks{School of Statistics and Data Science, Institute of Data Science and Statistics, Shanghai University of Finance and Economics, Shanghai 200433, China. E-mail: feng.xingdong@mail.shufe.edu.cn}
\and
Xinhong Jiang
\thanks{
School of Mathematics and Statistics, Wuhan University, Wuhan, 430072, China.
Email: xinhongjiang@whu.edu.cn
}
\and 
Yuling Jiao
\thanks{
School of Artificial Intelligence,
 and Hubei Key Laboratory of Computational Science, Wuhan University, Wuhan, 430072, China.
Email: yulingjiaomath@whu.edu.cn
} 
\and
Lican Kang
\thanks{Institute for Math and AI,   Hubei Key Laboratory of Computational Science, and
School of Artificial Intelligence,
Wuhan University, Wuhan, 430072, China. Email: kanglican@whu.edu.cn }
\and
Junwei Liu
\thanks{
School of Mathematics and Statistics, Wuhan University, Wuhan, 430072, China.
Email: liujunweimath@whu.edu.cn
}
}
  \maketitle
} \fi

\if0\anon
{
  \bigskip
  \bigskip
  \bigskip
  \begin{center}
    {\LARGE\bf \paperTitle}
  \end{center}
  \medskip
} \fi

\bigskip
\begin{abstract}
This work
proposes deep nonparametric  Instrumental variable quantile regression (IVQR), a two-stage   estimator that combines conditional diffusion modeling with a 
kernel-smoothed conditional moment formulation. In the first stage, we estimate the joint conditional distribution of the outcome and endogenous covariates given the instrument using a variance-preserving conditional diffusion model.  In the second stage, we approximate the conditional moment operator through Monte Carlo sampling and a kernel-smoothed surrogate for the indicator function, and then estimate the structural quantile function by empirical risk minimization over 
deep neural networks.
We establish an excess-risk bound for the proposed estimator and derive end-to-end total variation guarantees for the conditional diffusion model under unbounded support, explicitly accounting for score estimation, early stopping, and discretization errors.
Our theory is developed under a polynomial-tail envelope on the data
distribution and degenerates continuously to the
exponential  setting: as the tail index grows, the obtained 
excess-risk rate converges to the minimax-optimal rate of
nonparametric regression,
thus our heavy-tailed theory covers the classical light-tailed
nonparametric guarantees as a limiting case.
Simulation studies and a real-data application demonstrate that the proposed method outperforms existing nonparametric IVQR approaches, with gains that become increasingly pronounced as the dimensionality of the covariates and instruments increases.
\end{abstract}

\noindent%
\vfill

\newpage
\spacingset{1.8}  

\section{Introduction}

Endogeneity, the phenomenon in which an observed covariate is
correlated with the unobserved error term, is a pervasive obstacle
to causal inference across economics, epidemiology, and the social
sciences. In its presence, ordinary regression does not identify the
underlying structural relationship, and the instrumental variables
(IV) method is the canonical remedy: one exploits an exogenous
variable 
that affects the outcome  
only through the
endogenous covariate.   
Classical IV analysis, however,
targets only the conditional mean and therefore captures one
moment of the outcome distribution. 
IVQR
introduced by
\citet{chernozhukov2005iv,chernozhukov2006instrumental}, extends the
framework to the full conditional quantile function and so permits
the modelling of heterogeneous structural effects across the outcome
distribution. The IVQR model is given by 
\begin{align}\label{model1}
y = f_{0}(\mathbf{x}) + \epsilon,
\qquad
\mathbb{P}\!\bigl(\epsilon\leq 0\bigm|\mathbf{z}\bigr)=\tau,
\end{align}
where $y \in \mathbb{R}$ is the outcome, $\mathbf{x} \in \mathbb{R}^{d}$ is the endogenous covariate, $\mathbf{z} \in \mathbb{R}^{p}$ is the exogenous instrumental variable, $\epsilon \in \mathbb{R}$ is an unobserved error term that may be correlated with $\mathbf{x}$, $\tau \in (0,1)$ is the quantile level, and $f_{0} : \mathbb{R}^{d} \to \mathbb{R}$ is the unknown structural quantile function. The identifying restriction in 
 \eqref{model1} is
equivalent to the conditional moment equation
\begin{align}\label{eq:ivqr}
\mathbb{E}\!\bigl[\phi(y-f_{0}(\mathbf{x}))\bigm|\mathbf{z}\bigr]=0,
\qquad
\phi(u):=\tau-\mathbf{I}(u\leq 0),
\end{align}
or, equivalently, to an operator equation. Let $p_{\mathbf{x},y\mid\mathbf{z}}$ denote the conditional
density of $(\mathbf{x},y)$ given $\mathbf{z}$, and define the
operator
\begin{align}\label{eq:t}
\mathcal{T}f(\mathbf{z})\;:=\;
\iint \mathbf{I}(y-f(\mathbf{x})\leq 0)\,p_{\mathbf{x},y\mid\mathbf{z}}\,
d\mathbf{x}\,\mathrm{d}y,~ f:\mathbb{R}^d \rightarrow \mathbb{R}.
\end{align}
Then the identifying restriction can be written as $\mathcal{T}f_{0}(\mathbf{z})=\tau$.

Estimating the structural quantile function $f_{0}$ introduced above is a central problem in nonparametric IVQR, and a variety of methods have been developed for this purpose. Representative approaches
include the Tikhonov-regularized estimator
of \citet{horowitz2007nonparametric}, the penalized sieve minimum-distance
framework of \citet{chen2012estimation}, whose convergence rate matches
the optimal rate established for nonparametric mean IV regression by
\citet{hall2005nonparametric}, and the Tikhonov-regularized inverse estimator
of \citet{gagliardini2012nonparametric} with pointwise asymptotic normality,
all focus on regularizing the inversion of $\mathcal{T}$ and provide
well-understood convergence-rate and asymptotic-distribution guarantees.
However, the kernel-based and fixed-basis sieve approximations on which these
estimators rely are typically developed and analyzed under low-dimensional
settings of $\mathbf{x}$ and $\mathbf{z}$. For instance, the foundational
construction of \citet{horowitz2007nonparametric} is presented for a scalar
endogenous variable and a scalar instrument. As in classical nonparametric
estimation more broadly, these methods may suffer from performance degradation
as the dimensions of $\mathbf{x}$ and $\mathbf{z}$ increase, thereby limiting
their applicability in modern high-dimensional regimes.
Nevertheless, to the best of our knowledge, none of these nonparametric IVQR
estimators employs a deep neural network (DNN) function class as the hypothesis space,
thereby leaving the expressive power of modern deep architectures unexploited.
A primary technical obstacle is the discontinuity of the indicator-based moment
function $\phi(u)=\tau-\mathbf{I}(u\leq 0)$ in $f$, which complicates the use
of gradient-based optimization methods.
Recently, a parallel but {exogenous-covariate} literature has developed
deep nonparametric estimators of conditional quantiles using neural
networks \citep{padilla2022quantile,shen2021deep,feng2024deep,shen2024nonparametric},
some of which attain minimax-optimal rates under smoothness or
compositional assumptions on $f_{0}$ and suitable regularity conditions.
These methods, however, treat $\mathbf{x}$ as exogenous and identify
$f_{0}$ with the {observational} conditional quantile of $y$ given
$\mathbf{x}$, which differs from the {structural} quantile
in \eqref{model1} whenever endogeneity is present. Consequently, when
applied to an endogenous covariate $\mathbf{x}$, such approaches yield
inconsistent estimates of the structural target $f_{0}$ that motivates IVQR.

In this work, we propose deep nonparametric IVQR   
(DIVQR), a two-stage estimator that targets precisely this gap:
an endogeneity-aware deep nonparametric estimator of $f_{0}$ that
bypasses the low-dimensional sieve or Tikhonov approximation of
$\mathcal{T}$ used in classical nonparametric IVQR. The starting
point is that
$\mathcal{T}f(\mathbf{z})=
\mathbb{E}_{(\mathbf{x},y)\sim p_{\mathbf{x},y\mid\mathbf{z}}}
\bigl[\mathbf{I}(y-f(\mathbf{x})\leq 0)\bigr]$
is an expectation under the conditional density
$p_{\mathbf{x},y\mid\mathbf{z}}$. Rather than approximating this
expectation by a polynomial sieve projection of the moment
function 
or by a Tikhonov-regularized
inversion, we learn $p_{\mathbf{x},y\mid\mathbf{z}}$ itself.
Our choice to model $p_{\mathbf{x},y\mid\mathbf{z}}$ by a conditional diffusion model is motivated by both empirical
and theoretical considerations. Diffusion
models  
\citep{ho2020denoising,song2020score}
have emerged as the leading approach to high-dimensional generative
modelling: in vision, audio, and tabular benchmarks they routinely
match the data distribution at orders of magnitude higher dimension
and stronger multi-modality than is achievable by classical kernel
density estimators or copula models, which underpin the
sieve/Tikhonov approximations to $\mathcal{T}f$ used in
nonparametric IVQR. 
Specifically, a variance-preserving conditional diffusion model is
trained to approximate $p_{\mathbf{x},y\mid\mathbf{z}}$. Samples generated
from the learned model, together with a kernel smoothing of the indicator
function 
\citep{kaplan2017smoothed,kaplan2022smoothed},
are used to construct a Monte Carlo approximation of $\mathcal{T}f$. Finally, we construct a deep
nonparametric IV estimator by minimizing the resulting smoothed empirical
moment loss over a class of DNNs. We implement the two stages on independent samples, ensuring statistical
decoupling between generative modeling and moment optimization; see
Section \ref{sec:method} for details. Consequently, this constructed deep
nonparametric IV estimator inherits the identification of the IVQR
model \eqref{model1}, exploits the expressive power of deep networks,
and avoids both the low-dimensional sieve or Tikhonov regularization
of classical nonparametric IVQR and the non-differentiability of the
sample pinball objective.
Beyond the algorithmic development, we derive a non-asymptotic
excess risk bound for the proposed estimator, in which the error of the
conditional diffusion model enters explicitly, with a key technical
contribution in establishing its error bound. 
Specifically, existing convergence-rate
results for diffusion models typically assume compactly
supported data, using boundedness to control tail contributions in
the reverse-time
stochastic differential equation (SDE) \citep{chen2022distribution,chen2023score,oko2023diffusion,jiao2025model}.
We therefore establish an end-to-end total variation convergence
guarantee for the conditional diffusion model under unbounded
support of $(\mathbf{x},y,\mathbf{z})$, handling tails through a
polynomially decaying (heavy-tailed) envelope condition on
$p_{\mathbf{x},y\mid\mathbf{z}}$ and a sub-Gaussian assumption on
$\mathbf{z}$, and tracking every
source of error in the pipeline: score approximation by ReLU
networks, statistical estimation of the score, early stopping, and the
numerical discretization of the reverse-time SDE.
The polynomial-tail envelope is among the weakest tail regimes
considered for end-to-end diffusion guarantees, and our theory
degenerates continuously to the exponential (Gaussian-envelope)
setting: as the tail index grows, the
excess-risk rate converges to the minimax-optimal rate of
nonparametric  regression for H\"older-smooth functions,
and the first-stage total-variation rate recovers exactly its
Gaussian-envelope counterpart, thus the proposed heavy-tailed theory
covers the classical light-tailed nonparametric guarantees as a
limiting case.

\subsection{Contributions}

Our main contributions are as follows.
\begin{itemize}
\item 
We propose DIVQR, a two-stage estimator that approximates the
conditional moment operator $\mathcal{T}f$ in \eqref{eq:t} via a
Monte Carlo average based on samples drawn from a learned conditional
diffusion model, combined with a kernel-smoothing approximation of the
indicator function. To the best of our knowledge, no existing IVQR
estimator approximates $\mathcal{T}f$ using samples from a learned
conditional density; existing approaches instead rely on
finite-dimensional parametric forms, sieve projections,
Tikhonov-regularized inverses, or debiased machine-learning scores
for high-dimensional linear models. Furthermore, we conduct extensive
simulation studies and a real-data application that provide empirical
evidence for the effectiveness and practical utility of the proposed
method.

\item 
We derive a non-asymptotic total variation convergence rate for the
first-stage conditional diffusion model that tracks all sources of error
in the pipeline and holds under unbounded support of $(\mathbf{x}, y, \mathbf{z})$
with conditional densities decaying only polynomially, extending
prior compact-support and Gaussian-envelope analyses and making the
theory directly applicable to IVQR under realistic, heavy-tailed data. Propagating this error to the second stage yields a
non-asymptotic excess risk bound for the deep nonparametric IV estimator.
The polynomial-tail theory contains the classical light-tailed
guarantees as its boundary case: as the tail index $\alpha$ grows, the
excess-risk exponent increases to $2\beta_q/(d+2\beta_q)$, the
minimax-optimal rate of nonparametric regression for
$\beta_q$-H\"older functions, while the first-stage rate recovers
exactly the Gaussian-envelope total-variation rate of
\citet{fu2024unveil}; the price of polynomial tails is quantified by
two explicit factors that vanish in the light-tail limit.

\end{itemize}

\subsection{Notations}\label{subsec:notations}
We introduce the notations used throughout this paper.
The symbols $\mathbb{R}$ and $\mathbb{N}$ denote the
real numbers and the natural numbers, respectively, and for any
integer $N\geq 0$ we let $[N]:=\{0,1,\ldots,N\}$. The $\ell^{2}$-norm
of a vector
$\mathbf{x}=(x_{1},\ldots,x_{d})^{\top}\in\mathbb{R}^{d}$ is
$\Vert\mathbf{x}\Vert:=\sqrt{\sum_{i=1}^{d}x_{i}^{2}}$. For any
function $f:\mathbb{R}^{d}\to\mathbb{R}$, the symbol $\nabla f$
denotes its gradient. For a subset $K\subseteq\mathbb{R}^{d}$, the
$L^{\infty}(K)$-norm of $f$ is
$\Vert f\Vert_{L^{\infty}(K)}:=\sup_{\mathbf{x}\in K}|f(\mathbf{x})|$,
and for a vector-valued function
$\mathbf{v}:\mathbb{R}^{d}\to\mathbb{R}^{d}$,
$\Vert\mathbf{v}\Vert_{L^{\infty}(K)}
:=\sup_{\mathbf{x}\in K}\Vert\mathbf{v}(\mathbf{x})\Vert$.
The symbols $\mathbb{E}[\cdot]$ and $\mathbb{P}(\cdot)$ denote
expectation and probability, respectively. We write
$\mathbf{I}(\cdot)$ for the indicator function and use the same
boldface letter with a positive integer subscript,
$\mathbf{I}_{d}\in\mathbb{R}^{d\times d}$, for the identity matrix
of dimension $d$ (so that, e.g., $\mathbf{I}_{d+1}$ denotes the
$(d+1)\times(d+1)$ identity matrix).
For a real number $a$, we write $\lfloor a\rfloor$
for the largest integer no greater than $a$.
For two non-negative sequences $\{a_{n}\}$ and $\{b_{n}\}$, we
write $a_{n}\lesssim b_{n}$ or 
$a_{n} \gtrsim b_{n}$ 
if there exists a constant $C>0$ such
that $a_{n}\leq C b_{n}$ or $a_{n}\geq C b_{n}$ for all $n$ sufficiently large. The
asymptotic notation $f(\mathbf{x})=\mathcal{O}\bigl(g(\mathbf{x})\bigr)$
is used to indicate that $f(\mathbf{x})\leq C\,g(\mathbf{x})$ for
some constant $C>0$, and $\widetilde{\mathcal{O}}(\cdot)$ further
suppresses logarithmic factors in the asymptotic analysis. We
further write $a_{n}\asymp b_{n}$ if $a_{n}\lesssim b_{n}$ and
$a_{n}\gtrsim b_{n}$ both hold.

\subsection{Paper Organization}

The remainder of the paper is organized as follows.
Section~\ref{sec:method} formally introduces the proposed two-stage method for nonparametric IVQR.
Section~\ref{sec:theory}
provides non-asymptotic theoretical guarantees for the proposed method.
Section~\ref{sec:numerical} reports simulation studies  
and a real-data analysis.
Section~\ref{sec:conclusion} concludes the paper.
Proofs of all theoretical
results, additional numerical configurations, 
preliminary definitions,
and related work 
are given in the Supplemental Material.

\section{Method}\label{sec:method}

In this section, we present our two-stage estimator for the
structural quantile $f_{0}$ in the IVQR model~\eqref{model1}.
The
construction is driven by the operator equation
$\mathcal{T}f_{0}(\mathbf{z})=\tau$ in  
\eqref{eq:t},
which faces two computational obstacles in finite samples: the
indicator $\mathbf{I}(\cdot\leq 0)$ inside $\mathcal{T}$ is
discontinuous in $f$, and the conditional density
$p_{\mathbf{x},y\mid\mathbf{z}}$ that defines $\mathcal{T}$ is
unknown. We address these obstacles in two stages.

\noindent 
\textbf{Stage~I: Learning $p_{\mathbf{x},y\mid\mathbf{z}}$ by a
conditional diffusion model.}
We learn the conditional density
$p_{\mathbf{x},y\mid\mathbf{z}}$ by training a variance-preserving
conditional diffusion model whose forward dynamics gradually
transform a draw from $p_{\mathbf{x},y\mid\mathbf{z}}$ into an
approximately standard Gaussian, and whose reverse dynamics generate
new samples from the learned distribution conditional on
$\mathbf{z}$. The drift of the reverse-time SDE depends on the
conditional score $\nabla_{\mathbf{w}}\log
p_{t}(\mathbf{w}\mid\mathbf{z})$,
where
$\mathbf{w} := (\mathbf{x}, y)$ and $p_t$ denotes the density of the
forward process. 
With dataset
$\mathcal{D}_{1}
:=\bigl\{(\mathbf{x}_{i},y_{i},\mathbf{z}_{i})\bigr\}_{i=1}^{\widetilde{n}}
\overset{\mathrm{i.i.d.}}{\sim}p_{\mathbf{x},y,\mathbf{z}}
$, we estimate this score by minimizing a denoising score-matching
objective over a deep ReLU score network
$\mathbf{s}_{\boldsymbol\theta}(t,\mathbf{w};\mathbf{z})$. Numerical
simulation of the learned reverse SDE then yields, for every
instrument value $\mathbf{z}$ in the sample, an i.i.d.\ Monte Carlo
sample
$\{(\widehat{\mathbf{x}}_{i,\mathbf{z}},
\widehat{y}_{i,\mathbf{z}})\}_{i=1}^{\widetilde{m}}\sim
\widehat{p}_{\mathbf{x},y\mid\mathbf{z}}$, where $\widehat{p}_{\mathbf{x},y\mid\mathbf{z}}$, obtained from the conditional diffusion model, denotes the estimator of $p_{\mathbf{x},y\mid\mathbf{z}}$. 
The full Stage-I construction,
including the forward and
reverse SDEs, the score-matching loss, and the choice of
discretization, is presented in detail in Section \ref{subsec:cdm}.

\noindent 
\textbf{Stage~II: Smoothed empirical risk over a ReLU DNN class.}
With the Stage-I samples in hand, we construct a Monte Carlo approximation of the operator $\mathcal{T}$ by replacing the inner expectation with a sample average and the indicator with a kernel-smoothed surrogate. We then obtain a deep nonparametric IVQR estimator by minimizing the resulting 
least-squares objective over a ReLU DNN class.

Recall that in \eqref{eq:ivqr}, the indicator function is non-smooth.
To address this issue, we adopt a kernel-smoothing technique.
Such smoothing improves statistical accuracy and facilitates computation,
as established in  
\cite{kaplan2017smoothed,kaplan2022smoothed}.
Let $K(\cdot)$ be a kernel function, i.e., a nonnegative, symmetric map
$K:\mathbb{R}\to\mathbb{R}$ with $\int_{\mathbb{R}}K(v)\,\mathrm{d}v=1$,
assumed to be bounded and compactly supported on $[-1,1]$. For a bandwidth
parameter $h>0$, define $K_h(v):=h^{-1}K(v/h)$. Typical examples include
the uniform and triangular kernels. Define the smoothed indicator
\begin{align}\label{eqind}
 \mathbf{I}_h(v):= \int \mathbf{I}(u \leq 0) K_h(v - u) du
 =\int_{u \geq v/h} K(u) du
 =G(-v/h),
\end{align}
where the second equality follows from the change of variables
$s=(v-u)/h$, under which $\{u\le0\}=\{s\ge v/h\}$. 
And we also have 
$ G(u):=\int_{-\infty}^{u}K(s)\,\mathrm{d}s
$ by using the symmetry of $K$. 
In
particular, $\mathbf{I}_h(v)\to1$ as $v\to-\infty$ and
$\mathbf{I}_h(v)\to0$ as $v\to\infty$, matching the limits of
$\mathbf{I}(v\le0)$. Then
$\mathbf{I}_h$ is a continuous and differentiable surrogate of
$v\mapsto \mathbf{I}(v\leq 0)$, takes values in $[0,1]$, and satisfies
$\mathbf{I}_h(v)\to \mathbf{I}(v\leq 0)$ pointwise as $h\downarrow 0$.
Using $\mathbf{I}_h$, define the smoothed operator 
\begin{align}\label{eq:min}
\mathcal{T}_hf(\mathbf{z}):=
\iint
\mathbf{I}_h(y-f(\mathbf{x})) p_{\mathbf{x},y\mid\mathbf{z}}
d\mathbf{x}dy.
\end{align} 
Then $\mathcal{T}_h f \to \mathcal{T}f$ as $h\downarrow 0$, and in
particular $\mathcal{T}_h f_0(\mathbf{z}) \to \tau$ since
$\mathcal{T}f_0(\mathbf{z})=\tau$.
The operator $\mathcal{T}_h$ is not directly computable, as the conditional
density $p_{\mathbf{x},y\mid\mathbf{z}}$ is unknown. To address this,
we use the Stage-I conditional diffusion model to estimate this conditional distribution
and approximate the integral in \eqref{eq:min} via Monte Carlo sampling.
Specifically, given the estimator
$\widehat{p}_{\mathbf{x},y\mid\mathbf{z}}$ trained on $\mathcal{D}_{1}$,
we draw i.i.d.\ samples
$
(\widehat{\mathbf{x}}_{i,\mathbf{z}}, \widehat{y}_{i,\mathbf{z}})_{i=1}^{\widetilde{m}}
\sim \widehat{p}_{\mathbf{x},y\mid\mathbf{z}},
$
and define the Monte Carlo approximation
\begin{equation}\label{eq:def_That}
\widehat{\mathcal{T}}_{h}f(\mathbf{z})
:=
\frac{1}{\widetilde{m}}\sum_{i=1}^{\widetilde{m}}
\mathbf{I}_{h}\!\left(\widehat{y}_{i,\mathbf{z}}
- f(\widehat{\mathbf{x}}_{i,\mathbf{z}})\right).
\end{equation}
Using the independent sample
$
\mathcal{D}_{2}
:=
\bigl\{(\mathbf{x}_{j},y_{j},\mathbf{z}_{j})\bigr\}_{j=1}^{n}
\overset{\mathrm{i.i.d.}}{\sim} p_{\mathbf{x},y,\mathbf{z}},
$
we define the empirical loss
\begin{equation}\label{eq:def_Lhat}
\widehat{\mathcal{L}}_{\widehat{\mathcal{T}}_{h}}(f)
:=
\frac{1}{n}\sum_{j=1}^{n}
\bigl|\widehat{\mathcal{T}}_{h}f(\mathbf{z}_{j})-\tau\bigr|^{2}.
\end{equation}
The proposed estimator $\widehat{f}_{h}$ is then defined as
\begin{align}\label{eq:emin}
\widehat{f}_{h}
\in
\argmin_{f\in\mathcal{F}}
\widehat{\mathcal{L}}_{\widehat{\mathcal{T}}_{h}}(f),
\end{align}
where $\mathcal{F}$ is a ReLU DNN class.

In our method, we assume access to two independent i.i.d.\ samples,
$\mathcal{D}_{1}$ and $\mathcal{D}_{2}$, of sizes $\widetilde{n}$ and
$n$ respectively, drawn from
the same population $p_{\mathbf{x},y,\mathbf{z}}$, with
$\mathcal{D}_{1} \perp\!\!\!\perp \mathcal{D}_{2}$. The Stage-I sample
$\mathcal{D}_{1}$ 
is used to train the conditional diffusion model,
while the Stage-II sample $\mathcal{D}_{2}$ 
is used to construct the
smoothed empirical risk and the resulting least-squares estimator over
a ReLU DNN class. This sample-splitting protocol decouples
the randomness of the first-stage conditional diffusion estimator from
that of the second-stage empirical risk. In particular,
$\widehat{p}_{\mathbf{x},y\mid\mathbf{z}}$ can be treated as fixed when
analyzing the stochastic fluctuations of
$\widehat{\mathcal{L}}_{\widehat{\mathcal{T}}_{h}}$ on $\mathcal{D}_{2}$,
while the total variation guarantee of the conditional diffusion model
on $\mathcal{D}_{1}$ can be invoked independently of the Stage-II
optimization. The two sample sizes are decoupled deliberately: the
operator-estimation error is governed by the Stage-I size
$\widetilde{n}$ and the function-estimation error by the Stage-II size
$n$, and tuning $\widetilde{n}$ and $n$ separately delivers the optimal
rate, which is ultimately determined by $n$. For notational clarity, we use distinct indices
(e.g., $i$ and $j$) to denote observations from $\mathcal{D}_{1}$ and
$\mathcal{D}_{2}$, respectively.
The complete workflow 
is summarized in
Algorithm \ref{alg:two_stage_ivqr}.

\begin{algorithm}[H]
\caption{Deep nonparametric IVQR via Conditional Diffusion Models}
\label{alg:two_stage_ivqr}
\DontPrintSemicolon
\SetKwInput{KwIn}{Input}
\SetKwInput{KwOut}{Output}

\KwIn{
Two independent i.i.d.
samples  
$\mathcal{D}_{1}=\{(\mathbf{x}_{i},y_{i},\mathbf{z}_{i})\}_{i=1}^{\widetilde{n}}$
and
$\mathcal{D}_{2}=\{(\mathbf{x}_{j},y_{j},\mathbf{z}_{j})\}_{j=1}^{n}$;
quantile level $\tau\in(0,1)$;
kernel $K$ and bandwidth $h>0$;
Monte Carlo size $\widetilde{m}$.
}
\KwOut{Estimator $\widehat{f}_{h}\in\mathcal{F}$.}

\textbf{(I) Stage~I, Conditional diffusion training on $\mathcal{D}_{1}$.}
Train a conditional diffusion model on the Stage-I sample $\mathcal{D}_{1}$ to obtain
the conditional distribution $\widehat{p}_{\mathbf{x},y\mid\mathbf{z}}$
(see Section~\ref{subsec:cdm}).\;

\textbf{(II) Stage~II, Empirical risk minimization on $\mathcal{D}_{2}$.}\;
\For{$j=1,\dots,n$}{
Sample $\bigl\{\bigl(\widehat{\mathbf{x}}_{i,\mathbf{z}_{j}},\,
\widehat{y}_{i,\mathbf{z}_{j}}\bigr)\bigr\}_{i=1}^{\widetilde{m}}\sim
\widehat{p}_{\mathbf{x},y\mid\mathbf{z}_{j}}$ 
via the conditional diffusion model.
Compute $\widehat{\mathcal{T}}_{h}f(\mathbf{z}_{j})$ by~\eqref{eq:def_That}.\;
}
Form the empirical loss $\widehat{\mathcal{L}}_{\widehat{\mathcal{T}}_{h}}(f)$
via \eqref{eq:def_Lhat} and compute $\widehat{f}_{h}$ by \eqref{eq:emin}.\;

\Return{$\widehat{f}_{h}$.}
\end{algorithm}

\subsection{
Conditional Diffusion Model
}\label{subsec:cdm}

In this section, 
we consider the target distribution as the conditional distribution of 
$\mathbf{w}:=(\mathbf{x},y) \in \mathbb{R}^{d+1}$ given $\mathbf{z}\in \mathbb{R}^p$, denoted as 
$P_{\mathbf{w}|\mathbf{z}}(\cdot)$. Then, we formulate the conditional diffusion model over the time interval $t \in [0,T]$ to learn the conditional distribution.

\noindent
\textbf{Diffusion Process.}
The forward diffusion process is defined as:
\begin{equation}\label{sde: cond_OU_forward}
 \mathrm{d}\mathbf{W}_{t}^{F}=-\mathbf{W}_{t}^{F}\mathrm{d}t+\sqrt{2}\mathrm{d}\mathbf{B}_t, ~ \mathbf{W}_{0}^{F} \sim p_0(\mathbf{w}| \mathbf{z}).
\end{equation}
It determines a diffusion process, starting with a conditional distribution $p_0(\mathbf{w}| \mathbf{z})$ at time $t=0$, that approaches $\mathcal{N}(\mathbf{0},\mathbf{I}_{d+1})$ as $t\rightarrow\infty$. In this context, we set $p_0(\mathbf{w}| \mathbf{z})$ as the target conditional distribution $P_{\mathbf{w}|\mathbf{z}}(\cdot)$. Conditioning on some observation $\mathbf{z} \in \mathcal{Z}$, the transition probability distribution from $\mathbf{W}_0^F$ to $\mathbf{W}_t^F$ is given by $\mathbf{W}_t^F|\mathbf{W}_0^F\sim\mathcal{N}(m_t\mathbf{W}_0^F,\sigma_t^2\mathbf{I}_{d+1})$, where $m_t := e^{-t}$, $\sigma_t := \sqrt{1-e^{-2t}}$. 

The SDE above can be reversed if we know the score of the distribution
at each intermediate time step, $\nabla\log p_t(\mathbf{w}|\mathbf{z})$.
The classical time-reversal results of \citet{anderson1982reverse,haussmann1986time} imply that the reverse-time dynamics of the forward process~\eqref{sde: cond_OU_forward} again form a diffusion, and the conditional
reverse SDE reads
\begin{equation}\label{sde: reverse_cond_OU}
 \mathrm{d}\mathbf{W}_t^R=\left[-\mathbf{W}_t^R-2\nabla\log p_t(\mathbf{W}_t^R|\mathbf{z}) \right] \mathrm{d}t+\sqrt{2}\mathrm{d}\mathbf{B}_t^R,
\end{equation}
where $t$ flows from $t=\infty$ to $t=0$ and $\mathbf{B}_t^R$ is a time reverse Brownian motion. For convenience, we reformulate the reverse SDE \eqref{sde: reverse_cond_OU} in a forward version by switching time direction $t\rightarrow T - t$:
\begin{equation}\label{sde: forward_cond_OU}
\mathrm{d}\mathbf{W}_t=\left[\mathbf{W}_t+2\nabla\log p_{T-t}(\mathbf{W}_t|\mathbf{z}) \right] \mathrm{d}t+\sqrt{2}\mathrm{d}\mathbf{B}_t,~\mathbf{W}_0\sim p_{T}(\mathbf{w}|\mathbf{z}),~\mathbf{W}_T\sim p_{0}(\mathbf{w}|\mathbf{z}).
\end{equation}
It is obvious that $\mathbf{W}_t\sim p_{T-t}(\mathbf{w}|\mathbf{z})$.

\noindent
\textbf{Score Matching.}
We apply denoising score-matching
techniques \citep{hyvarinen2005estimation,vincent2011connection}
to estimate the target conditional score function of SDE
\eqref{sde: forward_cond_OU}, defined as $\mathbf{b}^{*}(t,\mathbf{w},\mathbf{z}) := \nabla\log p_t(\mathbf{w}|\mathbf{z})$. It can be verified that $\mathbf{b}^{*}$ minimizes the loss function $\mathcal{L}(\mathbf{b})$ over all measurable functions, where
\begin{equation*}
 \mathcal{L}(\mathbf{b}) := \frac{1}{T-T_0} \int_{T_0}^{T} \mathbb{E}_{(\mathbf{z},\mathbf{W}_t^F)}\Vert\mathbf{b}(t,\mathbf{W}_t^F,\mathbf{z})-\nabla\log p_t(\mathbf{W}_t^F|\mathbf{z})\Vert^2 \mathrm{d}t,
\end{equation*}
where $0 < T_0 < T < \infty$. We notice that the selection of $T_0$ is motivated by the necessity to preclude the score function from exhibiting a blow-up at $t=0$, concomitantly facilitating the stabilization of the training process for the model.
Using denoising score matching, we can equivalently represent this objective as:
\begin{align*}
 \mathcal{L}(\mathbf{b})&=\frac{1}{T-T_0} \int_{T_0}^{T} \mathbb{E}_{(\mathbf{z},\mathbf{W}_0^F)}\mathbb{E}_{\mathbf{W}_t^F|(\mathbf{W}_0^F,\mathbf{z})}\left\Vert\mathbf{b}(t,\mathbf{W}_t^F,\mathbf{z})+\frac{\mathbf{W}_t^F-m_t\mathbf{W}_0^F}{\sigma_t^2}\right\Vert^2 \mathrm{d}t\\
 &=\frac{1}{T-T_0} \int_{T_0}^{T} \mathbb{E}_{(\mathbf{z},\mathbf{W}_0^F)}\mathbb{E}_{\mathcal{N}(\mathbf{U};\mathbf{0},\mathbf{I}_{d+1})}\left\Vert\mathbf{b}(t,m_t\mathbf{W}_0^F+\sigma_t\mathbf{U},\mathbf{z})+\frac{\mathbf{U}}{\sigma_t}\right\Vert^2 \mathrm{d}t.
\end{align*}
Given $\widetilde{n}$ i.i.d.\ samples $\left\{(\mathbf{W}^F_{0,i}, \mathbf{z}_{i})\right\}_{i=1}^{\widetilde{n}}$ from $p(\mathbf{z})p_0(\mathbf{w}|\mathbf{z})$, and $m$ i.i.d. samples $\left\{(t_j, \mathbf{U}_j)\right\}_{j=1}^{m}$ from $U[T_0, T]$ and $\mcN(0, \mI_{d+1})$, we can utilize the empirical risk minimizer (ERM) as the estimator of the conditional score function $\mathbf{b}^{*}$. This ERM estimator $\wh{\mathbf{b}}$ is defined as
\begin{equation}\label{escore}
 \wh{\mathbf{b}}\in \mathop{\arg\min}_{\mathbf{b}\in\mrN}\wh{\mcL}(\mathbf{b}) := \frac{1}{m\widetilde{n}}\sum_{j=1}^{m}\sum_{i=1}^{\widetilde{n}}\left\Vert\mathbf{b}(t_j,m_{t_j}\mathbf{W}_{0,i}^F+\sigma_{t_j}\mathbf{U}_j,\mathbf{z}_i)+\frac{\mathbf{U}_j}{\sigma_{t_j}}\right\Vert^2,
\end{equation}
where the optimization is performed over the score network class
$\mathrm{NN}(L,M,J,\kappa)$ (shorthand for $\mathrm{NN}$), namely a class of ReLU DNN.
Specifically, the conditional score
$\mathbf{b}^{\ast}(t,\mathbf{w},\mathbf{z})
=\nabla_{\mathbf{w}}\log p_{t}(\mathbf{w}\mid\mathbf{z})$, viewed as a vector
field on
$[T_{0},T]\times\mathbb{R}^{d+1}\times\mathbb{R}^{p}$ taking values in
$\mathbb{R}^{d+1}$, is estimated over the ReLU DNN class
$
\mathrm{NN}(L,M,J,\kappa)
\;:=\;
\mathcal{F}_{p+d+2,\,d+1}(L,M,J,\kappa),
$
i.e., $\mathcal{F}_{d_1,d_2}(L,M,J,\kappa)$ specialized to input
dimension $d_{1}=p+d+2$ (concatenating the scalar diffusion time, the
$(d+1)$-dimensional state, and the $p$-dimensional instrument) and
output dimension $d_{2}=d+1$.

\noindent
\textbf{Euler--Maruyama Discretization.} Given the estimated score function $\wh{\mathbf{b}}$, as defined in \eqref{escore}, we can formulate an SDE initializing from the prior distribution:
\begin{equation}\label{sampling_SDE}
 \mrd\wh{\mathbf{W}}_t=[\wh{\mathbf{W}}_t+2\wh{\mathbf{b}}(T-t,\wh{\mathbf{W}}_t,\mathbf{z})]\mrd t+\sqrt{2}\mrd \mathbf{B}_t,~\wh{\mathbf{W}}_0\sim \mcN(\mathbf{0},\mathbf{I}_{d+1}).
\end{equation}
Now, we employ a discrete-time approximation for the sampling dynamics \eqref{sampling_SDE}. 
Let 
$$0=t_0<t_1<\cdots<t_K=T-T_0,~ K \in \mathbb{N}^+,$$ 
be the discretization points on $[0,T-T_0]$. We consider the sampling scheme:
\begin{equation}\label{EM_scheme}
 \mrd\wt{\mathbf{W}}_t=[\wt{\mathbf{W}}_{t_i}+2\wh{\mathbf{b}}(T-t_i,\wt{\mathbf{W}}_{t_i},\mathbf{z})]\mrd t+\sqrt{2}\mrd \mathbf{B}_t,~\wt{\mathbf{W}}_{t_0}\sim \mcN(\mathbf{0},\mathbf{I}_{d+1}),~t\in[t_i,t_{i+1}),
\end{equation}
for $i=0,1,\cdots,K-1$. 
Subsequently, we can utilize dynamics \eqref{EM_scheme} to generate new samples.
The proposed conditional diffusion model is summarized as an algorithm in the Supplementary Material.

\section{Theory}\label{sec:theory}

In this section, we develop theoretical guarantees for the proposed
two-stage estimator. Our primary objective is to bound the excess risk
$$
\mathcal{L}_{\mathcal{T}}(\widehat{f}_h) - \mathcal{L}_{\mathcal{T}}(f_0),
~\mbox{with}~
\mathcal{L}_{\mathcal{T}}(f)
:=
\mathbb{E}_{\mathbf{z}}
\bigl|\mathcal{T}f(\mathbf{z}) - \tau\bigr|^2.
$$
Moreover, we also have
$\mathbb{E}_{\mathbf{z}}
\bigl|\mathcal{T}f(\mathbf{z}) - \mathcal{T}f_0(\mathbf{z})\bigr|^2=
\mathcal{L}_{\mathcal{T}}(f) - \mathcal{L}_{\mathcal{T}}(f_0)
$ for each measurable $f$ since
$\mathcal{T}f_0(\mathbf{z})=\tau$.
We then introduce the population risk associated with the smoothed operator,
$
\mathcal{L}_{\mathcal{T}_h}(f)
:=
\mathbb{E}_{\mathbf{z}}
\bigl|\mathcal{T}_h f(\mathbf{z}) - \tau\bigr|^2,
$
and let
\begin{align}\label{eq:fh}
f_h \in \arg\min_{f} \mathcal{L}_{\mathcal{T}_h}(f),
\end{align}
denote the corresponding population minimizer.
The proposed method consists of a sequence of approximation and
estimation steps, each contributing to the overall excess risk. In
Stage {I}, a conditional diffusion model is used to learn the
conditional distribution $p_{\mathbf{x},y\mid\mathbf{z}}$, giving rise
to a distribution learning error. In Stage {II}, the
non-smooth indicator function is approximated via kernel smoothing,
which introduces a bias through the discrepancy between the smoothed
operator $\mathcal{T}_h$ and the target operator $\mathcal{T}$.
Finally, the structural quantile function is estimated via empirical
risk minimization over a class of ReLU neural networks, leading to an 
 estimation error.
A central difficulty lies in controlling these error components
simultaneously and quantifying their propagation across the two-stage
procedure. Our analysis develops a unified non-asymptotic framework in
which the excess risk admits a decomposition into (i) conditional
distribution learning error, (ii) smoothing bias, and (iii) estimation 
error from empirical risk minimization. This decomposition enables a
fully end-to-end characterization of the estimator and forms the basis
for establishing explicit convergence rates in
Theorem~\ref{theorem:excess_risk_heavy_tail}.
Our theory is developed for heavy-tailed conditional laws: the
conditional density of $\mathbf{w}=(\mathbf{x},y)$ given $\mathbf{z}$ is
only required to decay polynomially in $\|\mathbf{w}\|$. We now
introduce the assumptions under which this analysis is carried out.
\begin{assumption}\label{ass:kernel}
$K(\cdot)$ is a bounded and nonnegative kernel function and is
compactly supported on $[-1,1]$.
\end{assumption}

\begin{assumption}\label{ass:hfh}
The structural quantile function $f_0$ in \eqref{model1} and the
smoothed quantile function $f_h$ in \eqref{eq:fh} both belong to the
H\"older class $\mathcal{H}^{\beta_q}(\mathbb{R}^{d},B)$ with smoothness
 index $\beta_q\ge1$.
\end{assumption}

\begin{assumption} 
\label{ass:poly_decay}
Let $C_{p}$ and $C_{\ell}$ be two positive constants and let
$f\in\mathcal{H}^{\beta_f}(\mathbb{R}^{d+1}\times\mathbb{R}^{p},B)$ be a
H\"older factor with smoothness index $\beta_f\geq1$.
We assume $f(\mathbf{w},\mathbf{z})\geq C_{\ell}$ for all
$(\mathbf{w},\mathbf{z})$, and that the conditional density of
$\mathbf{w}=(\mathbf{x},y)$ given $\mathbf{z}$ admits the
representation
\[
p(\mathbf{w}\mid\mathbf{z})
=C_{p}\,\bigl(1+\|\mathbf{w}\|_{2}^{2}\bigr)^{-\alpha/2}\,
f(\mathbf{w},\mathbf{z}),
\]
where the tail index satisfies $\alpha>d+3$. 
\end{assumption}

\begin{assumption}\label{ass:subgauss_z_poly}
The marginal density of
the instrument $\mathbf{z}$ has sub-Gaussian tails:
$p(\mathbf{z})\leq\exp\!\bigl(-C_{\mathcal{Z}}\|\mathbf{z}\|^{2}/2\bigr)$ for some constant $C_{\mathcal{Z}}>0$.
\end{assumption}

\begin{remark}
\label{rem:assumptions}
Assumption~\ref{ass:kernel} is standard for smoothed estimating
equations and smoothed IVQR  
\citep{kaplan2017smoothed,kaplan2022smoothed};
admissible choices include the Epanechnikov, biweight and triweight
kernels. Assumption \ref{ass:hfh} imposes standard smoothness conditions
on the target functions $f_0$ and $f_h$.

In Assumption~\ref{ass:poly_decay}, the conditional density is only
required to decay polynomially,
$p(\mathbf{w}\mid\mathbf{z})\asymp\|\mathbf{w}\|_2^{-\alpha}$ as
$\|\mathbf{w}\|_2\to\infty$. The envelope
governs the tails of $\mathbf{w}\mid\mathbf{z}$, while the
H\"older-smooth factor $f\in\mathcal{H}^{\beta_f}$ determines the
smoothness exponent that drives the convergence rate; the lower bound
$f\ge C_{\ell}$ prevents degeneracy of the score
$\nabla_{\mathbf{w}}\log p_{t}(\mathbf{w}|\mathbf{z})$ along the forward
Ornstein--Uhlenbeck flow, although the diffused density itself admits no
uniform lower bound in $\mathbf{w}$.

In Assumption~\ref{ass:subgauss_z_poly}, the sub-Gaussian condition on
$\mathbf{z}$ relaxes the bounded-instrument assumptions of classical
smoothed IVQR     
\citep{kaplan2017smoothed}
and is satisfied by
Gaussian, sub-Gaussian and bounded instruments. Only the
instrument is required to be light-tailed; the endogenous pair
$(\mathbf{x},y)$ may be arbitrarily heavy-tailed within the polynomial
family of Assumption~\ref{ass:poly_decay}. Together,
Assumptions~\ref{ass:poly_decay} and~\ref{ass:subgauss_z_poly} replace
the bounded-support requirement pervasive in the diffusion-model
literature \citep{chen2023score,jiao2025model,oko2023diffusion} 
and the
smoothed IVQR literature 
\citep{kaplan2017smoothed,kaplan2022smoothed}
with
tail conditions weak enough to accommodate power-law data.
\end{remark}

\begin{remark}[Anatomy of the tail condition $\alpha>d+3$]
\label{rem:tail_index}
The single restriction $\alpha>d+3$ in
Assumption~\ref{ass:poly_decay} is exactly calibrated to the moment
structure of the polynomial envelope. Since
$p(\mathbf{w}\mid\mathbf{z})\asymp\|\mathbf{w}\|_2^{-\alpha}$ with
$\mathbf{w}\in\mathbb{R}^{d+1}$, the radial computation gives the
moment equivalence
\begin{align*}
\mathbb{E}\bigl[\|\mathbf{w}\|_2^{k}\,\big|\,\mathbf{z}\bigr]
\;\asymp\;\int_1^{\infty}r^{\,k+d-\alpha}\,\mathrm{d}r<\infty
\quad\Longleftrightarrow\quad
k<\alpha-(d+1),
\end{align*}
so $\alpha>d+3$ is equivalent to a finite conditional second moment,
$\mathbb{E}[\|\mathbf{w}\|_2^{2}\mid\mathbf{z}]<\infty$, while all
moments of order $k\ge\alpha-(d+1)$ may be infinite. The two stages of
our analysis consume this condition at different strengths, and we
record the precise entry points.\\
\noindent
\textbf{Stage I (conditional diffusion model).}
The full strength $\alpha>d+3$
enters the Stage-I analysis at two binding points, from which its
remaining occurrences in the schedule follow algebraically. First, the drift of the reverse-time dynamics
has a finite second moment along the forward marginals,
$\sup_{t\in[0,T]}\mathbb{E}_{p_t}[\|\mathbf{w}\|^2\mid\mathbf{z}]
\le\overline{M}_2<\infty$,   established in the Supplementary Material,
whose Beta-integral
$B\bigl(\tfrac{d+3}2,\tfrac{\alpha-d-3}2\bigr)$ is finite if and only
if $\alpha>d+3$; this is the earliest quantitative use of
$\alpha>d+3$ and feeds the path-increment and localization bounds of
the Girsanov argument (the accompanying Fisher-information term is
tail-free). Second, the Stage-I
inflation factor
$\Phi=1+\frac{3\beta_f+2}{\alpha-d-3}$ of \eqref{eq:gamma_main} is
finite and exceeds $1$ exactly on $\alpha>d+3$; equivalently, the
grid exponent $a_1=\frac{3\beta_f+2}{\alpha-d-1+3\beta_f}$ fixed in the
Stage-I schedule   satisfies $a_1<1$ if and only if
$\alpha>d+3$, which is what drives the local-polynomial grid error
$\delta_N=\mathcal{O}((R/N)^{\beta_f})$ to zero; here $N$ is the
per-coordinate number of grid cells and $R$ the local-polynomial grid
radius fixed in the Stage-I schedule,
related by $R\asymp N^{a_1}$ up to a $\sqrt{\log N}$ factor,
so that
$R/N=\mathcal{O}(N^{a_1-1}\sqrt{\log N})\to0$ precisely when $a_1<1$.\\
\noindent
\textbf{Stage II (excess risk).} 
The Stage-II analysis is strictly less
demanding: as detailed in Remark~\ref{rem:stageII_tail}, it uses only
$\alpha>d$ for the envelope integrability behind
Lemmas~\ref{lemma:bound_I}--\ref{lemma:Smothness_Error}
and the operator-Lipschitz bound established in the Supplementary Material,
and $\alpha>d+1$ for the covariate tail
bound, which is equivalent to
$\psi>0$ in \eqref{eq:psi_main}. Both are implied by $\alpha>d+3$, so
the single condition of Assumption~\ref{ass:poly_decay} governs the
entire two-stage procedure, with no strengthening required anywhere.

Moreover, a detailed comparison with existing tail assumptions in the literature is provided in the Supplementary Material.
\end{remark}

We next introduce several additional quantities.   Let $\psi\in(0,1)$ be the Stage-II approximation slowdown factor, i.e., 
\begin{equation}\label{eq:psi_main}
\psi
:=\frac{\alpha-d-1}{\alpha-d-1+2\beta_q}
\;\in(0,1).
\end{equation}
$\psi>0$ is equivalent to $\alpha>d+1$, which is already implied by the
tail condition $\alpha>d+3$ of Assumption~\ref{ass:poly_decay}; in
particular, no strengthening of Assumption~\ref{ass:poly_decay} is
required for the Stage-II analysis (see
Remarks~\ref{rem:tail_index} and~\ref{rem:stageII_tail}).
Moreover, $\gamma\in(0,\tfrac12)$ is
the Stage-I exponent
\begin{equation}\label{eq:gamma_main}
\gamma=\frac{\beta_f}{2\beta_f+(p+d+1)\Phi},
\qquad
\Phi=\frac{\alpha-d-1+3\beta_f}{\alpha-d-3}
=1+\frac{3\beta_f+2}{\alpha-d-3}.
\end{equation}

We now present our main result in the
following theorem, which establishes the convergence rate of the
excess risk for the deep nonparametric IVQR estimator.
\begin{theorem} 
\label{theorem:excess_risk_heavy_tail}
Suppose Assumptions~\ref{ass:kernel}, \ref{ass:hfh},
\ref{ass:poly_decay} and \ref{ass:subgauss_z_poly} hold. Let $\psi\in(0,1)$,
$\gamma\in(0,\tfrac12)$, and $\Phi>1$ be as defined in \eqref{eq:psi_main} and
\eqref{eq:gamma_main}. Set the kernel bandwidth
and the Monte-Carlo sample size as
$n^{-1}\le h\le n^{-\frac{2\beta_q\psi}{d+2\beta_q\psi}}$
and
$\widetilde m\gtrsim n^{\frac{4\beta_q\psi}{d+2\beta_q\psi}}$,
respectively. Assume that the Stage-I and Stage-II sample sizes  
satisfy $\widetilde{n}\gtrsim
n^{\frac{2\beta_q\psi\,(2\beta_f+(p+d+1)\Phi)}{\beta_f\,(d+2\beta_q\psi)}}$,
and that the Stage-I time--noise sample size satisfies 
$m\ge\widetilde{n}^{2+4\gamma}$.
The
Stage-I score estimator $\widehat{\mathbf{b}}$ in~\eqref{escore}
is structured with
\begin{gather*}
L=\mathcal{O}(\log^{4}\widetilde{n}),\qquad
M=\mathcal{O}\!\bigl(\widetilde{n}^{\frac{(p+d+1)\Phi}{2\beta_f+(p+d+1)\Phi}}(\log\widetilde n)^{d+8}\bigr),\\
J=\mathcal{O}\!\bigl(\widetilde{n}^{\frac{(p+d+1)\Phi}{2\beta_f+(p+d+1)\Phi}}(\log\widetilde n)^{d+10}\bigr),\qquad
\kappa=\exp\!\bigl(\mathcal{O}(\log^{4}\widetilde{n})\bigr),
\end{gather*}
and the deep nonparametric IVQR estimator $\widehat{f}_{h}$ in
\eqref{eq:emin} is structured with
\begin{equation*}
L=\mathcal{O}(\log n),\
M=\mathcal{O}\!\bigl(n^{\frac{d}{d+2\beta_q\psi}}\log n\bigr),\
J=\mathcal{O}\!\bigl(n^{\frac{d}{d+2\beta_q\psi}}\log n\bigr),\
\kappa=\mathcal{O}\!\bigl(n^{\frac{\max(d,\beta_q)}{d+2\beta_q\psi}}\bigr).
\end{equation*}
Then,
\[
\mathbb{E}_{\mathcal{D},\mathcal{T},\mathcal{U},\mathcal{Z}}
\mathbb{E}_{\mathbf{z}}
\Bigl[\bigl|\mathcal{T}\widehat f_h(\mathbf{z})
-\mathcal{T}f_0(\mathbf{z})\bigr|^2\Bigr]
=\widetilde{\mathcal{O}}\!\Bigl(n^{-\frac{2\beta_q\psi}{d+2\beta_q\psi}}\Bigr).
\]
\end{theorem}

\begin{remark}[Rate interpretation and the role of $\psi$]\label{rem:main_rate}
Write $\gamma_{\mathrm{ex}}:=\frac{2\beta_q\psi}{d+2\beta_q\psi}$ for
the exponent in Theorem~\ref{theorem:excess_risk_heavy_tail}. The
factor $\psi$ is the exact exchange rate between network resolution
and clipping radius under a polynomial covariate tail: by
Lemma~\ref{lemma:approx_error_poly_2}, the Stage-II network of width
parameter $N_2$ clipped at radius $R$ incurs the on-box error
$(R/N_2)^{2\beta_q}$ and the off-box tail error $R^{-(\alpha-1-d)}$,
and equating the two exponents at $R=N_2^{1-\psi}$ gives
$(R/N_2)^{2\beta_q}=R^{-(\alpha-1-d)}=N_2^{-2\beta_q\psi}$ with $\psi$
as in \eqref{eq:psi_main} (defined precisely in the Supplementary Material). In particular,
$\psi>0$ is equivalent to $\alpha>d+1$, already implied by
Assumption~\ref{ass:poly_decay} (Remarks~\ref{rem:tail_index}
and~\ref{rem:stageII_tail}), and $\psi\uparrow1$ as $\alpha\to\infty$,
so that
$\gamma_{\mathrm{ex}}\uparrow\frac{2\beta_q}{d+2\beta_q}$, the minimax
exponent for $\beta_q$-H\"older functions in dimension $d$: the
light-tailed theory is recovered as a limiting case. The same
multiplicative structure appears in recent heavy-tailed diffusion
theory: for unconditional sampling under the one-sided envelope
$p_0(\cdot)\le C(1+\|\cdot\|_2^2)^{-(1+\gamma'+d')/2}$ on $\mathbb{R}^{d'}$
with Sobolev smoothness $\beta'$ (primed, as are $d'$ and $\gamma'$,
to distinguish their parameters from ours), \citet{yu2026heavy}
obtain the total variation rate
$\widetilde{\mathcal{O}}(n^{-\kappa'})$,
$\kappa'=\frac{2\beta'(\gamma'+1)}
{4\beta'(d'+\gamma'+1)+d'(d'+2(\gamma'+1))}$,
with $\kappa'\uparrow\frac{\beta'}{2\beta'+d'}$, the
density-estimation minimax exponent, as $\gamma'\to\infty$.

Furthermore, a detailed comparison with existing theoretical results for nonparametric IVQR is provided in the Supplementary Material.
\end{remark}

\begin{remark}[Stage decoupling and tuning-parameter choices]\label{rem:decoupling}
The final rate is determined entirely by the Stage-II nonparametric
problem in dimension $d$ with smoothness $\beta_q$; the Stage-I
conditional diffusion model enters only through the sample-size
coupling
$\widetilde{n}\gtrsim
n^{\frac{2\beta_q\psi\,(2\beta_f+(p+d+1)\Phi)}{\beta_f\,(d+2\beta_q\psi)}}
=n^{\gamma_{\mathrm{ex}}/\gamma}$,
under which the Stage-I total-variation error $\widetilde n^{-\gamma}$
(see Supplementary Material)
is
$\mathcal{O}(n^{-\gamma_{\mathrm{ex}}})$ and does not limit the
overall rate.
The tail assumption affects the Stage-I convergence exponent only through $\Phi$. As $\alpha\to\infty$,
and hence $\Phi\downarrow1$,  
the Stage-I total-variation error $\widetilde n^{-\gamma}$
recovers the
rate
$\widetilde{\mathcal O}\bigl(\widetilde n^{-\beta_f/(p+d+1+2\beta_f)}\bigr)$
of conditional diffusion under a Gaussian envelope with a
H\"older-smooth factor, the framework of \citet{fu2024unveil}. The
pair $(\Phi,\psi)$ thus quantifies the entire statistical price of
polynomial tails, and this price vanishes continuously in the
light-tail limit. Finally, the bandwidth $h$ and the Monte Carlo size
$\widetilde m$ are chosen so that the smoothing bias $\mathcal O(h)$
and the fluctuation $\widetilde m^{-1/2}$ are negligible relative to
$n^{-\gamma_{\mathrm{ex}}}$.
\end{remark}

\subsection{
Proof sketch of Theorem \ref{theorem:excess_risk_heavy_tail}
}\label{sec:ps}
To analyze the excess risk, we perform the following decomposition:
\begin{align*}
&\mathbb{E}_{\mathcal{D},\mathcal{T},\mathcal{U},\mathcal{Z}}
[\mathcal{L}_{\mathcal{T}}(\widehat{f}_h)]-
\mathcal{L}_{\mathcal{T}}(f_0)\\
=&\mathbb{E}_{\mathcal{D},\mathcal{T},\mathcal{U},\mathcal{Z}}
[\mathcal{L}_{\mathcal{T}}(\widehat{f}_h)
-\mathcal{L}_{\mathcal{T}_h}(\widehat{f}_h)]
+\mathbb{E}_{\mathcal{D},\mathcal{T},\mathcal{U},\mathcal{Z}}[\mathcal{L}_{\mathcal{T}_h}(\widehat{f}_h)]
-\mathcal{L}_{\mathcal{T}}(f_0)\\
=&\underbrace{\mathbb{E}_{\mathcal{D},\mathcal{T},\mathcal{U},\mathcal{Z}}
[\mathcal{L}_{\mathcal{T}}(\widehat{f}_h)
-\mathcal{L}_{\mathcal{T}_h}(\widehat{f}_h)]}_{\mbox{Term}~ \textbf{I}}+\underbrace{\mathbb{E}_{\mathcal{D},\mathcal{T},\mathcal{U},\mathcal{Z}}\left[\mathcal{L}_{\mathcal{T}_h}(\widehat{f}_h)-\mathcal{L}_{\mathcal{T}_h}(f_h)\right]}_{\mbox{Term}~ \textbf{II}}+\underbrace{\mathcal{L}_{\mathcal{T}_h}(f_h)
-\mathcal{L}_{\mathcal{T}}(f_0)}_{\mbox{Term}~ \textbf{III}}.
\end{align*}
Consequently, it suffices to control the three terms above.
Term \textbf{I} captures the discrepancy between the target operator
$\mathcal{T}$ and its smoothed counterpart $\mathcal{T}_h$, while
Term \textbf{III} represents the corresponding approximation bias at the
population level. Both terms are governed by the smoothing error.
Term \textbf{II} corresponds to the estimation error arising from
empirical risk minimization and incorporates the impact of conditional
distribution learning. 
The convergence analysis of the conditional diffusion model used to control this latter component is provided in the Supplementary Material.

\noindent
\textbf{Bound Term I and Term III.} 
Term \textbf{I} represents the excess risk error induced by replacing
the operator $\mathcal{T}$ with its smoothed counterpart
$\mathcal{T}_h$. Under Assumption \ref{ass:kernel}, kernel
smoothing arguments imply that this discrepancy is controlled by the
bandwidth parameter $h$; the only distributional input is a uniform
bound on the conditional density of $y-\widehat{f}_h(\mathbf{x})$ given
$\mathbf{z}$, which under Assumption~\ref{ass:poly_decay} follows from
the integrability of the polynomial envelope in the $\mathbf{x}$
direction, this is where the requirement $\alpha>d$ enters, and we
verify it explicitly in the Supplementary Material.
For Term \textbf{III}, we further decompose
\begin{align*}
\mathcal{L}_{\mathcal{T}_h}(f_h)
-
\mathcal{L}_{\mathcal{T}}(f_0)
=
\mathcal{L}_{\mathcal{T}_h}(f_h)
-
\mathcal{L}_{\mathcal{T}_h}(f_0)
+
\mathcal{L}_{\mathcal{T}_h}(f_0)
-
\mathcal{L}_{\mathcal{T}}(f_0)
\le
\mathcal{L}_{\mathcal{T}_h}(f_0)
-
\mathcal{L}_{\mathcal{T}}(f_0),
\end{align*}
where the inequality follows from the optimality of $f_h$ as the
minimizer of $\mathcal{L}_{\mathcal{T}_h}$. Consequently, Term
\textbf{III} is also governed by the discrepancy between
$\mathcal{T}_h$ and $\mathcal{T}$ and is therefore controlled by $h$
under Assumption \ref{ass:kernel}.
The precise upper bounds for Terms \textbf{I} and \textbf{III} are
established in the following Lemmas
\ref{lemma:bound_I}--\ref{lemma:Smothness_Error}.

\begin{lemma}[Bound Term \textbf{I}]\label{lemma:bound_I}
Under Assumptions~\ref{ass:kernel} and~\ref{ass:poly_decay}, it holds that
\begin{align*}
\mathbb{E}_{\mathcal{D},\mathcal{T},\mathcal{U},\mathcal{Z}}
\bigl[\mathcal{L}_{\mathcal{T}}(\widehat{f}_h)
-\mathcal{L}_{\mathcal{T}_h}(\widehat{f}_h)\bigr]
\ \lesssim\ h.
\end{align*}
\end{lemma}


\begin{lemma}[Bound Term \textbf{III}]\label{lemma:Smothness_Error}
Under Assumptions~\ref{ass:kernel} and~\ref{ass:poly_decay}, it holds that
\begin{align*}
\mathcal{L}_{\mathcal{T}_h}(f_h)
-\mathcal{L}_{\mathcal{T}}(f_0)
\lesssim h.
\end{align*}
\end{lemma}


\noindent
\textbf{Bound Term II.}
We first make an error decomposition for Term \textbf{II}.
Recall that
\[
\textbf{II}
=
\mathbb{E}_{\mathcal{D},\mathcal{T},\mathcal{U},\mathcal{Z}}
\Big[\mathcal{L}_{\mathcal{T}_h}(\widehat{f}_h)-\mathcal{L}_{\mathcal{T}_h}(f_h)\Big],~
\mbox{with}
~
\mathcal{L}_{\mathcal{T}_h}(f)=\mathbb{E}_{\mathbf{z}}\big|\mathcal{T}_h f(\mathbf{z})-\tau\big|^2.
\]
By introducing the sample-based operator $\widehat{\mathcal{T}}_h$,
we can decompose Term \textbf{II} into
\begin{equation}\label{eq:II_decomp}
\mathcal{L}_{\mathcal{T}_h}(\widehat{f}_h)-\mathcal{L}_{\mathcal{T}_h}(f_h)
=
\underbrace{\Big(\mathcal{L}_{\mathcal{T}_h}(\widehat{f}_h)-\mathcal{L}_{\widehat{\mathcal{T}}_h}(\widehat{f}_h)\Big)}_{\mathrm{\bf II\text{-}A}}
+
\underbrace{\Big(\mathcal{L}_{\widehat{\mathcal{T}}_h}(\widehat{f}_h)-\mathcal{L}_{\mathcal{T}_h}(f_h)\Big)}_{\mathrm{\bf II\text{-}B}
}.
\end{equation}
Term $\mathrm{\bf II\text{-}A}$ captures the operator error $\mathcal{T}_h-\widehat{\mathcal{T}}_h$ evaluated at $\widehat{f}_h$,
whereas $\mathrm{\bf II\text{-}B}$ captures the estimation error induced by minimizing an empirical criterion under $\widehat{\mathcal{T}}_h$.
Now, we bound these two terms as follows.

\medskip
\noindent\textbf{(1) Bound $\mathrm{\bf II\text{-}A}$.}
For any $f\in\mathcal{F}$,
\[
\mathcal{L}_{\mathcal{T}_h}(f)-\mathcal{L}_{\widehat{\mathcal{T}}_h}(f)
=
\mathbb{E}_{\mathbf{z}}
\Big[
\big(\mathcal{T}_h f(\mathbf{z})-\widehat{\mathcal{T}}_h f(\mathbf{z})\big)
\big(\mathcal{T}_h f(\mathbf{z})+\widehat{\mathcal{T}}_h f(\mathbf{z})-2\tau\big)
\Big]
\ \lesssim\
\mathbb{E}_{\mathbf{z}}\big|\mathcal{T}_h f(\mathbf{z})-\widehat{\mathcal{T}}_h f(\mathbf{z})\big|.
\]
Moreover,
\begin{align*}
\mathcal{T}_h f(\mathbf{z})-\widehat{\mathcal{T}}_h f(\mathbf{z})
=
&\iint \mathbf{I}_h(y-f(\mathbf{x}))\big(p_{\mathbf{x},y\mid\mathbf{z}}-\widehat{p}_{\mathbf{x},y\mid\mathbf{z}}\big)\,\mathrm{d}\mathbf{x}dy\\
&+\Big(\mathbb{E}_{\widehat{p}_{\mathbf{x},y\mid\mathbf{z}}}\mathbf{I}_h(y-f(\mathbf{x}))
-\frac{1}{\widetilde{m}}\sum_{i=1}^{\widetilde{m}} \mathbf{I}_h(\widehat{y}_{i,\mathbf{z}}-f(\widehat{\mathbf{x}}_{i,\mathbf{z}}))\Big).
\end{align*}
On the right-hand side of the above equality, the first part is controlled by the total variation distance
$\mathrm{TV}(\widehat{p}_{\mathbf{x},y\mid\mathbf{z}},p_{\mathbf{x},y\mid\mathbf{z}})$
since the kernel-smoothed indicator $\mathbf{I}_h$ is uniformly bounded.
Consequently, the theoretical analysis of the conditional diffusion model provided in the Supplementary Material yields the rate
$
\widetilde{n}^{-\gamma}
$
up to a factor $(\log\widetilde n)^{(d+20)/2}$, with $\gamma$ as in \eqref{eq:gamma_main}.
The second part is a bounded empirical fluctuation and contributes $\mathcal{O}(\widetilde{m}^{-1/2})$, which is dominated.
Hence,
\[
\mathbb{E}\big[|\mathrm{\bf II\text{-}A}|\big]
\ \lesssim\
\widetilde{n}^{-\gamma}\,(\log\widetilde n)^{(d+20)/2}+\frac{1}{\sqrt{\widetilde{m}}}.
\]

\medskip
\noindent\textbf{(2) Bound $\mathrm{\bf II\text{-}B}$.}
Define the empirical risk under operator $\mathcal{T}_h$ by
\[
\widehat{\mathcal{L}}_{\mathcal{T}_h}(f)
:=\frac{1}{n}\sum_{j=1}^n\big|\mathcal{T}_h f(\mathbf{z}_j)-\tau\big|^2,
\]
so that
$\mathbb{E}\big[\widehat{\mathcal{L}}_{\mathcal{T}_h}(f)\big]=\mathcal{L}_{\mathcal{T}_h}(f).
$
Introducing $\widehat{\mathcal{L}}_{\widehat{\mathcal{T}}_h}$ and a comparator $f\in\mathcal{F}$, we obtain the decomposition
\[
\mathrm{\bf II\text{-}B}
\ \le\
\underbrace{\big(\mathcal{L}_{\widehat{\mathcal{T}}_h}(\widehat{f}_h)-\widehat{\mathcal{L}}_{\widehat{\mathcal{T}}_h}(\widehat{f}_h)\big)}_{\mathbf{C}:~\text{generalization under }\widehat{\mathcal{T}}_h}
+
\underbrace{\big(\widehat{\mathcal{L}}_{\widehat{\mathcal{T}}_h}(f)-\widehat{\mathcal{L}}_{\mathcal{T}_h}(f)\big)}_{\mathbf{D}:~\text{operator mismatch on }\mathcal{D}_2}
+
\underbrace{\big(\widehat{\mathcal{L}}_{\mathcal{T}_h}(f)-\mathcal{L}_{\mathcal{T}_h}(f_h)\big)}_{\mathbf{E}:~\text{approximation~+~smoothing}}.
\]
It remains to bound the three terms above, which we analyze in turn below.
\begin{itemize}
\item
Term $\mathbf{C}$ measures the generalization gap of $\widehat{f}_h$
under the surrogate operator $\widehat{\mathcal{T}}_h$. Its analysis
must account for four error sources. The first source is the
stochastic fluctuation of the empirical risk on the Stage-{II}
sample $\mathcal{D}_2$, which we control through a symmetrization
argument followed by a $\delta$-covering of $\mathcal{F}$ at the
bandwidth-adapted scale
$\delta=h/n$ and a Bernstein-type concentration inequality, producing
a covering-complexity term of order
$\log\mathcal{N}(h/n,\mathcal{F},\|\cdot\|_{\infty})/n$; the Stage-II
clipping radius $R$ is polynomially large in $n$,
but $R$ enters the covering bound only
through $\log R=\mathcal{O}(\log n)$, so the complexity term retains
the form $N_2^{d}\,(\log n)^3/n$. The second source is the
Stage-{I} distributional learning error: the conditional distribution
$\widehat{p}_{\mathbf{x},y\mid\mathbf{z}}$ learned by the diffusion
model differs from the true $p_{\mathbf{x},y\mid\mathbf{z}}$. The third source is the
Monte Carlo fluctuation arising from approximating
$\mathcal{T}_h f$ by the empirical average over $\widetilde{m}$
samples drawn from $\widehat{p}_{\mathbf{x},y\mid\mathbf{z}}$, which
contributes a term of order $\widetilde{m}^{-1/2}$. The fourth source
is the coupling between $\widehat{f}_h$ and Term~{\bf II} itself:
because $\widehat{f}_h$ is the empirical risk minimizer of
$\widehat{\mathcal{L}}_{\widehat{\mathcal{T}}_h}$, controlling
$\mathbf{C}$ directly leads to a recursive inequality in which
$\mathbb{E}[\mathbf{C}]$ appears on both sides.
Consequently, it follows that
 \begin{align*}
\mathbb{E}_{\mathcal{D},\mathcal{T},\mathcal{U},\mathcal{Z}}[\mathbf{C}]&\lesssim\mathbb{E}_{\mathcal{D},\mathcal{T},\mathcal{U},\mathcal{Z}} [\mathbf{D}]+\mathbb{E}_{\mathcal{D},\mathcal{T},\mathcal{U},\mathcal{Z}} [\mathbf{E}]+ \mathbb{E}_{\mathcal{D},\mathcal{T},\mathcal{U},\mathcal{Z}}\mathbb{E}_{\mathbf{z}}\!\left[\mathrm{TV}(\widehat{p}_{\mathbf{x},y\mid\mathbf{z}},p_{\mathbf{x},y\mid\mathbf{z}})\right]\\&\quad+\frac{1}{\sqrt{\widetilde{m}}}+\frac{\log \mathcal{N}(h/n,\mathcal{F},\|\cdot\|_{\infty})}{n}+\frac{1}{n}+h.
\end{align*}
The conditional distribution learning error
$\mathbb{E}_{\mathcal{D},\mathcal{T},\mathcal{U},\mathcal{Z}}\mathbb{E}_{\mathbf{z}}\!\left[\mathrm{TV}(\widehat{p}_{\mathbf{x},y\mid\mathbf{z}},p_{\mathbf{x},y\mid\mathbf{z}})\right]$
is analyzed in the Supplementary Material,
while
$\mathbb{E}[\mathbf{D}]$ and $\mathbb{E}[\mathbf{E}]$ are controlled
in the subsequent paragraphs.

\item
Term $\mathbf{D}$ captures the discrepancy between the empirical
operators $\widehat{\mathcal{T}}_h$ and $\mathcal{T}_h$, which arises
from both conditional distribution learning and Monte Carlo
approximation. Specifically, we have
\begin{align*}
\mathbb{E}_{\mathcal{D},\mathcal{T},\mathcal{U},\mathcal{Z}}
\left[
\mathbf{D}
\right]
&\lesssim
\mathbb{E}_{\mathcal{D},\mathcal{T},\mathcal{U},\mathcal{Z}}\mathbb{E}_\mathbf{z}[\mathrm{TV}(\widehat{p}_{\mathbf{x},y\mid\mathbf{z}},p_{\mathbf{x},y\mid\mathbf{z}})]
+\frac{1}{\sqrt{\widetilde{m}}}.
\end{align*}
The first term reflects the conditional distribution learning error,
while the second term corresponds to the Monte Carlo fluctuation.

\item
In expectation over $\mathcal{D}_2$, Term $\mathbf{E}$ reduces to
$\mathcal{L}_{\mathcal{T}_h}(f)-\mathcal{L}_{\mathcal{T}_h}(f_h)$ and
reflects the intrinsic
approximation power of the ReLU DNN class $\mathcal{F}$ together with
the kernel smoothing bias. Its analysis splits into two
parts. The first part, the approximation error of $f_h$ over the
ReLU DNN class $\mathcal{F}$, is controlled by converting operator
distances into $\mathbf{x}$-marginal distances 
and then invoking
Lemma \ref{lemma:approx_error_poly_2} below, which constructs an
explicit ReLU network with clipped input and provides a quantitative
bound depending on the network parameters $(L_2,M_2,J_2,\kappa_2)$, the
truncation radius $R$, and the approximation level $\epsilon$. The
covariate marginal has only a polynomial tail,
so the truncation remainder
decays as $R^{-(\alpha-1-d)}$ and forces a polynomially large clipping
radius.
The second part, the smoothing bias of $\mathcal{T}_h$ at $f_h$, is
controlled by Lemma \ref{lemma:Smothness_Error} and is of order $h$.
By combining these two parts,
we can bound Term $\mathbf{E}$.
\end{itemize}

\begin{lemma}[Approximation error of the Stage-II network]
\label{lemma:approx_error_poly_2}
Suppose that Assumptions~\ref{ass:hfh}, \ref{ass:poly_decay} and~\ref{ass:subgauss_z_poly}
hold, given $N_2\gg1$, any $\epsilon\in(0,1]$ and any radius $R\ge1$, there exists a ReLU DNN $\mathrm{b}_{3}\in\mathrm{NN}(L_2,M_2,J_2,\kappa_2)$ such that
\[
\mathrm{b}_{3}(\mathbf{x})
=\mathrm{b}_{3}\!\left(\mathrm{b}_{\mathrm{clip}}(\mathbf{x},-R,R)\right),
\]
with structure
$L_2= \mathcal{O}\!\left(\log(1/\epsilon)+\log R\right)$,
$M_2=J_2= \mathcal{O}\!\left(N_2^{d}(\log(1/\epsilon)+\log R)\right)$, and
$\kappa_2= \mathcal{O}\!\left(\max\{BN_2^{d},\,BR^{r_q}/\epsilon,\,R\}\right)$,
where $r_q=\lceil\beta_q\rceil-1$.
Moreover,
\begin{align*}
\inf_{f\in\mathcal{F}}\mathbb{E}_{\mathbf{x}}|f(\mathbf{x})-f_{h}(\mathbf{x})|^{2}
&\le\mathbb{E}_{\mathbf{x}}|\mathrm{b}_{3}(\mathbf{x})-f_{h}(\mathbf{x})|^{2}
\lesssim
\bigl((R/N_2)^{\beta_q}+\epsilon\bigr)^{2}
+R^{-(\alpha-1-d)}.
\end{align*}
\end{lemma}
The
off-box error is the polynomial remainder
$R^{-(\alpha-1-d)}$, balanced against the on-box
error $(R/N_2)^{2\beta_q}$ by letting $R$ grow polynomially with $N_2$:
taking $R=N_2^{1-\psi}$ with $\psi$ as in \eqref{eq:psi_main}
equalises the two exponents, since
$(1-\psi)(\alpha-d-1)=2\beta_q\psi$, and
produces the exponent $2\beta_q\psi$ appearing in
Theorem~\ref{theorem:excess_risk_heavy_tail}. Since the off-box remainder carries no
network-size factor, the truncation cost is a pure tail probability,
and $\psi>0$ requires only $\alpha>d+1$.

\noindent\textbf{Summary.}
Combining \eqref{eq:II_decomp} with the bounds on
$\mathrm{\bf II\text{-}A}, \mathbf{C},\mathbf{D},\mathbf{E}$ above yields the claimed control of
Term~\textbf{II} in Lemma~\ref{lemma:theorem_excess_risk_poly}.


\begin{lemma}[Stage-II excess risk under polynomial decay]\label{lemma:theorem_excess_risk_poly}
Suppose that Assumptions~\ref{ass:kernel}, \ref{ass:hfh}, \ref{ass:poly_decay} and~\ref{ass:subgauss_z_poly}
hold. Let $\mathcal{F}=\mathcal{F}_{d,1}(L_2,M_2,J_2,\kappa_2)$ be the
Stage-II network class with the structure of
Lemma~\ref{lemma:approx_error_poly_2}, and let $\widehat f_h$ be the
empirical minimiser \eqref{eq:emin} over $\mathcal{F}$ on the Stage-II
sample of size $n$.
Then, given $N_2\gg1$, for any
$\epsilon\in(0,1]$ and any clipping radius $R\ge1$, we have
\begin{align*}
\mathbb{E}_{\mathcal{D},\mathcal{T},\mathcal{U},\mathcal{Z}} \left[\mathcal{L}_{\mathcal{T}_h}(\widehat{f}_h)-\mathcal{L}_{\mathcal{T}_h}(f_h)\right]
\lesssim 
&  \ \widetilde{n}^{-\gamma}\,(\log\widetilde n)^{(d+20)/2}+\frac{N_2^d(\log^2\epsilon^{-1}+\log^2R)\log(\epsilon^{-1}N_2Rnh^{-1})}{n}
\\
&+h+\left((R/N_2)^{\beta_q}+\epsilon\right)^2+ R^{-(\alpha-1-d)}+\frac{1}{\sqrt{\widetilde{m}}},
\end{align*}
where $\gamma$ is the Stage-I exponent of \eqref{eq:gamma_main}.
In particular, writing
$\psi=\tfrac{\alpha-d-1}{\alpha-d-1+2\beta_q}\in(0,1)$ for the Stage-II
slowdown factor of \eqref{eq:psi_main}, the Stage-II parameter choices
are
\[
N_2=\bigl\lfloor n^{\frac{1}{d+2\beta_q\psi}}\bigr\rfloor,
\qquad
R=N_2^{\,1-\psi}=N_2^{\frac{2\beta_q}{\alpha-d-1+2\beta_q}},
\qquad
\epsilon=(R/N_2)^{\beta_q}=N_2^{-\beta_q\psi}.
\]
Here,  $N_2$ is the per-coordinate resolution of the Stage-II network;
the clipping radius grows polynomially in $N_2$ at the exponent
$1-\psi$, chosen so that the on-box exponent $2\beta_q$ and the off-box
tail exponent $\alpha-d-1$ coincide, i.e.\
$(1-\psi)(\alpha-d-1)=2\beta_q\psi$; and $\epsilon$ is matched to the
on-box accuracy $(R/N_2)^{\beta_q}=N_2^{-\beta_q\psi}$. The excess-risk
rate resulting from these choices is derived in the proof of
Theorem~\ref{theorem:excess_risk_heavy_tail}.
\end{lemma}

\begin{remark}[Tail requirements of the Stage-II analysis]\label{rem:stageII_tail}
The proof of Lemma~\ref{lemma:theorem_excess_risk_poly}
invokes the tail index $\alpha$ of
Assumption~\ref{ass:poly_decay} at exactly two points, and both are
weaker than $\alpha>d+3$. First, $\alpha>d$ is needed so that the
envelope $(1+\|\mathbf{x}\|_2^2)^{-\alpha/2}$ is integrable over
$\mathbb{R}^d$; this integrability 
is the only distributional input to
the kernel-bias bounds of
Lemmas~\ref{lemma:bound_I}--\ref{lemma:Smothness_Error}
and to the operator-level Lipschitz bound,
which converts operator distances into $\mathbf{x}$-marginal distances
without any inverse-bandwidth factor. Second, $\alpha>d+1$ is needed
so that the covariate marginal obeys the tail bound
$\mathbb{P}(\|\mathbf{x}\|_\infty>R)\lesssim R^{-(\alpha-1-d)}$,
which is the sole source of the truncation term $R^{-(\alpha-1-d)}$ in
Lemma~\ref{lemma:approx_error_poly_2}; equivalently,
$\alpha>d+1$ is exactly the condition $\psi>0$ in
\eqref{eq:psi_main}. No moment of $y$ beyond the density envelope is
used at any point of the Stage-II argument. Consequently, the
Stage-II analysis holds under a strictly weaker tail condition than
the Stage-I diffusion analysis, which requires the full
$\alpha>d+3$ of Assumption~\ref{ass:poly_decay}, 
and the single condition
$\alpha>d+3$ therefore suffices for the entire two-stage procedure.
\end{remark}

\section{Numerical Studies}\label{sec:numerical}

In this section, we evaluate the proposed two-stage estimator
$\widehat{f}_{h}$ on both controlled simulation experiments and a
real-data application, with several goals in mind: to verify the
finite-sample behavior, 
to assess the estimator's robustness to distributional features 
commonly encountered in applied IVQR settings, 
and to demonstrate its practical usefulness in an empirical application.

\subsection{Simulation Studies}\label{sec:numerical:sim}

We evaluate the proposed estimator in the nonparametric IVQR model \eqref{model1}.
We focus on quantile levels $\tau\in\{0.25,0.50,0.75\}$ in the setting
$d=p=10$, with $f_{0}$ depending on all coordinates of $\mathbf{x}$
in a nonlinear, non-additive manner
(see~\eqref{eq:sim_f0_dim10}); two error designs, a Gaussian and a
heavy-tailed $t_{3}$ design, are considered to assess robustness.
Our implementation follows exactly the two-stage procedure developed in Section \ref{sec:method}:
we first learn the conditional law of $\mathbf{w}:=(\mathbf{x},y) \in \mathbb{R}^{d+1}$ given $\mathbf{z}\in \mathbb{R}^p$ by the
conditional diffusion model, 
and then estimate the structural
quantile function by the smoothed empirical risk minimization procedure in
Algorithm~\ref{alg:two_stage_ivqr}.
Performance is measured by the test mean squared error
$\mathrm{MSE}_{\mathrm{test}}(\widehat{f}_{h})$ in~\eqref{eq:sim_test_mse},
tracked along an evenly spaced grid of total sample sizes
$N_{\mathrm{total}}\in\{1{,}000,\dots,3{,}000\}$ under a $75\%/25\%$
train/test split, and the bandwidth $h$ is selected from a candidate
grid by minimizing the conditional-moment validation criterion.

Further implementation details and a complete description of the data-generating process are provided in the Supplementary Material.
We consider Gaussian and $t_3$ disturbances, with the results for the Gaussian design reported in the Supplementary Material.
We consider the following
$10$-dimensional structural regression function: 
\begin{equation}
f_0^{(10)}(x)
=
\frac12\sum_{r=0}^{1}
\left[
0.8\sin\!\bigl(x_{5r+1}+0.8x_{5r+2}\bigr)
+
0.05\,x_{5r+3}\cos(x_{5r+3})
+
0.2\,
\frac{x_{5r+4}x_{5r+5}}{1+x_{5r+4}^2+x_{5r+5}^2}
\right],
\label{eq:sim_f0_dim10}
\end{equation}
for \(x=(x_1,\dots,x_{10})^\top\in\mathbb{R}^{10}\).

\noindent
\textbf{Evaluation Metric.}
We evaluate the estimator on an independent test set
through the test mean squared error
\begin{equation}
\mathrm{MSE}_{\mathrm{test}}(\widehat f_h)
=
\frac{1}{n_{\mathrm{test}}}
\sum_{i=1}^{n_{\mathrm{test}}}
\bigl(\widehat f_h(\mathbf{x}_i^{\mathrm{te}})-f_0(\mathbf{x}_i^{\mathrm{te}})\bigr)^2.
\label{eq:sim_test_mse}
\end{equation}
The grid of total sample sizes is
$$
N_{\mathrm{total}}
\in
\{1000,1200,1400,1600,1800,2000,2200,2400,2600,2800,3000\}.
$$
For each \(N_{\mathrm{total}}\), we use the proportional split
$
n_{\mathrm{train}}=0.75\,N_{\mathrm{total}},
~
n_{\mathrm{test}}=0.25\,N_{\mathrm{total}}.
$

\begin{figure}[H]
\centering
\includegraphics[width=0.9\linewidth]{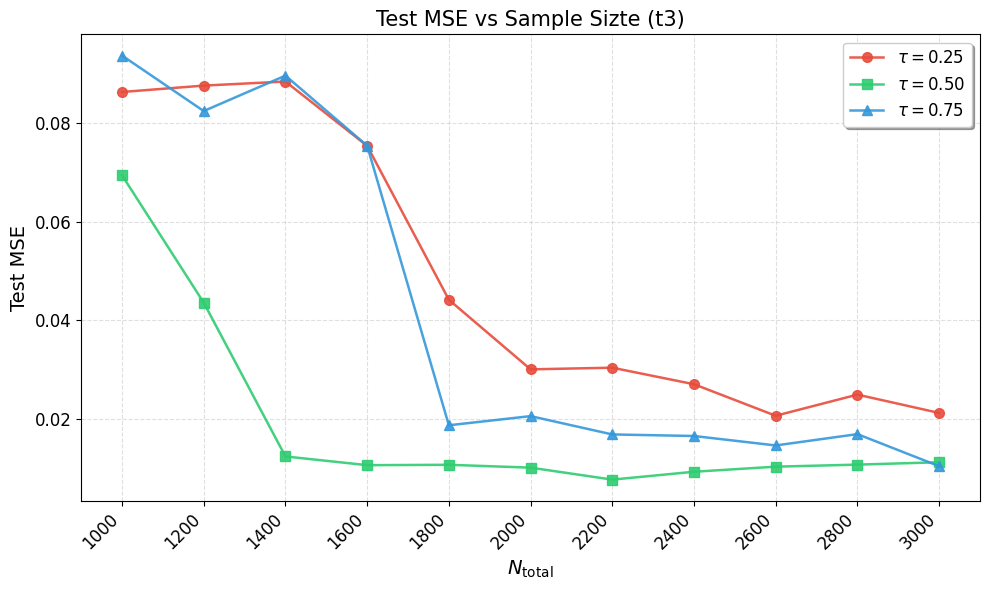}
 \caption{\(t_3\) design.}
 \label{fig:sim_t3_dim10_new}

\end{figure}

\noindent
\textbf{Simulation Results.}
Figure~\ref{fig:sim_t3_dim10_new} and
Table~\ref{tab:sim_dim10_summary}
summarize the test MSE of the proposed estimator under the 
\(t_3\)
design.

For \(\tau=0.25\), the test MSE decreases from \(0.0864\) at
\(N_{\mathrm{total}}=1000\) to \(0.0211\) at \(N_{\mathrm{total}}=3000\), with the
main reduction occurring after \(N_{\mathrm{total}}\approx 1600\).
For \(\tau=0.50\), the estimator again performs best, with the test MSE decreasing
from \(0.0694\) to \(0.0111\), and staying close to the \(10^{-2}\) level for
moderate and large sample sizes.
For \(\tau=0.75\), the test MSE decreases from \(0.0938\) to \(0.0103\), and a sharp
drop is observed between \(N_{\mathrm{total}}=1600\) and \(N_{\mathrm{total}}=1800\).
Hence, although the \(t_3\) design is more challenging in the low-sample regime due
to heavier tails, the proposed estimator still enters a stable low-error regime once
the sample size is sufficiently large.

The curves are not perfectly monotone at every adjacent
grid point. This is natural in the present setting, since each point corresponds to
one realization of a stochastic two-stage procedure involving score estimation,
Monte Carlo sampling, and neural-network optimization. Nevertheless, in both the
Gaussian and \(t_3\) designs, the overall trend is a substantial reduction of the
test MSE as \(N_{\mathrm{total}}\) increases. This is fully consistent with the
theoretical analysis in Section \ref{sec:theory}: as the sample size grows, the learned conditional
distribution \(\widehat p_{x,y\mid z}\) becomes more accurate, which improves the
approximation of the smoothed operator \(\mathcal{T}_h\) and yields a more accurate estimator
of the structural quantile function.

\begin{table}[H]
\centering
\caption{Test MSE of the proposed estimator at \(N_{\mathrm{total}}=1000,3000\) under the   \(t_3\) design.}
\label{tab:sim_dim10_summary}
\small
\renewcommand{\arraystretch}{1}
\begin{tabular*}{\textwidth}{@{\extracolsep{\fill}}lcccc}
\toprule
Design & \(N_{\mathrm{total}}\) & \(\tau=0.25\) & \(\tau=0.50\) & \(\tau=0.75\) \\
\midrule
\(t_3\) & 1000 & 0.0864 & 0.0694 & 0.0938 \\
\(t_3\) & 3000 & 0.0211 & 0.0111 & 0.0103 \\
\bottomrule
\end{tabular*}
\end{table}

\subsection{Real Data Analysis}\label{sec:numerical:realdata}

We evaluate the proposed estimator on a large-scale empirical
application in which the structural quantile function
\[
\mathbb{P}\!\left(\,y-f_{0}(\mathbf{x}) \leq 0 \,\bigm|\, \mathbf{z}\,\right)
\;=\;\tau,
\qquad
\tau\in\{0.25,\,0.50,\,0.75\},
\]
is estimated over an industry-level panel of U.S.\ manufacturing
activity, where output is modelled as a structural quantile function
of capital, labour, and energy inputs. Endogeneity arises because
input use is correlated with unobserved productivity shocks that also
drive output; the heterogeneous returns across the output
distribution, driven by sectoral composition and
capacity-utilization differences, cannot be recovered by an IV
mean regression. To address this, we construct a
five-dimensional continuous instrument from predetermined first-year
input cost shares interacted with the contemporaneous and
one-period-lagged log-changes of national input price indices, and
benchmark DIVQR against three nonparametric IVQR competitors that
share an identical Stage-{II} architecture but differ in how
$\mathcal{T}_{h}f(\mathbf{z})$ is estimated: a polynomial-sieve projection \citep{newey1997convergence},
a Nadaraya--Watson local-averaging estimator \citep{nadaraya1964estimating,watson1964smooth},
and a kernel-ridge regression \citep{caponnetto2007optimal}. Out-of-sample performance is reported by
the pinball loss
$\mathcal{P}_{\tau}(\widehat{f}_{h})$.

We establish the validity of the instrument design by verifying the linear identification conditions implied by the moment restriction. Details of this verification, along with descriptions of the benchmark methods, including Poly-Sieve IVQR, Nadaraya-Watson IVQR, and Kernel-Ridge IVQR, are provided in the Supplementary Material.

\noindent
\textbf{Dataset and Variable Construction.}
The data are drawn from the NBER--CES Manufacturing Industry
Database \citep{becker2021nberces}, a public industry-aggregate panel
jointly maintained by the National Bureau of Economic Research and the
U.S.~Census Bureau that covers all six-digit 1997-NAICS manufacturing
industries from 1958 through 2018. After dropping industry-years with non-positive inputs or outputs, the sample contains
\(n=26{,}339\) observations, randomly partitioned with a fixed seed into a training
set of size \(n_{\mathrm{train}}=19{,}754\) and a test set of size
\(n_{\mathrm{test}}=6{,}585\), corresponding to an approximate \(75\%/25\%\)
train--test split. 

All variables used below are constructed from the raw U.S.\ NBER--CES series.
The outcome is log real value-added
\begin{equation*}
y \;=\; \log\!\bigl(\mathrm{vship}/\pi_{Y}\,-\,\mathrm{matcost}/\pi_{M}\bigr),
\end{equation*}
where \(\mathrm{vship}\) is nominal shipments, \(\mathrm{matcost}\) is
nominal material cost, and \(\pi_{Y},\pi_{M}\) are the industry-specific
output and material price indices. The endogenous covariate vector is
\begin{equation*}
\mathbf{x}=(x_{1},x_{2},x_{3})^{\top}\in\mathbb{R}^{3},
\qquad
x_{1}=\log L,\quad
x_{2}=\log K,\quad
x_{3}=\log E,
\end{equation*}
with \(L\) production-worker hours, \(K\) real capital stock, and
\(E\) real energy consumption. All three components of \(\mathbf{x}\)
are endogenous in the structural equation, since industries select the
input bundle \((L,K,E)\) after observing unmodelled industry-level
productivity shocks that simultaneously drive \(y\).

To correct for this endogeneity we construct a five-dimensional
continuous instrument vector \(\mathbf{z}\in\mathbb{R}^{5}\) from
predetermined cost shares and national input price indices. Let
\(s_{J,i}^{(0)}\) denote the cost share of input
\(J\in\{M,E,K\}\) in the first year that industry \(i\) enters the
panel, and let \(\Delta\log\pi_{J,t}\) denote the annual log change in
the cross-industry mean of the national price index of input~\(J\).
We define
\begin{equation*}
\begin{aligned}
z_{1}&=s_{M,i}^{(0)}\,\Delta\log\pi_{M,t},
&\quad z_{2}&=s_{E,i}^{(0)}\,\Delta\log\pi_{E,t},
&\quad z_{3}&=s_{K,i}^{(0)}\,\Delta\log\pi_{K,t},
\\[0.25em]
z_{4}&=s_{M,i}^{(0)}\,\Delta\log\pi_{M,t-1},
&\quad z_{5}&=s_{E,i}^{(0)}\,\Delta\log\pi_{E,t-1}.&&
\end{aligned}
\end{equation*}
The shares \(s_{J,i}^{(0)}\) are fixed at the first year the industry
enters the panel and are therefore predetermined relative to any
industry-specific shock in year~\(t\); the national price indices
\(\pi_{J,t}\) vary only across time. The instruments \(z_{4},z_{5}\)
are lagged versions that reinforce the predetermination argument.

\noindent
\textbf
{Estimator.}
Our implementation of the two-stage procedure is
 identical to that used in the simulation. 
The smoothing function \(\mathbf{I}_{h}\) is the
uniform-kernel indicator with the bandwidth fixed at
\(h=0.1\).

\noindent
\textbf
{Evaluation Protocol.}
Because the structural function \(f_{0}\) is unknown in this
application, we evaluate \(\widehat{f}_{h}\) by out-of-sample
{pinball loss}. Given the test residual
\(e_{i}:=y_{i}^{\mathrm{te}}-\widehat{f}_{h}(\mathbf{x}_{i}^{\mathrm{te}})\),
the pinball loss of \(\widehat{f}_{h}\) at level \(\tau\) is
\begin{equation*}
\mathcal{P}_{\tau}(\widehat{f}_{h})
\;:=\;
\frac{1}{n_{\mathrm{test}}}
\sum_{i=1}^{n_{\mathrm{test}}}
\rho_{\tau}(e_{i}),
\qquad
\rho_{\tau}(u)\;:=\;\max\{\tau\,u,\,(\tau-1)\,u\}.
\end{equation*}
The population minimizer of \(\mathbb{E}\!\bigl[\rho_{\tau}(y-f(\mathbf{x}))\bigr]\)
over measurable functions is the observational conditional
\(\tau\)-quantile \(q_{\tau}(\mathbf{x})\) of \(y\) given
\(\mathbf{x}\), which coincides with the structural quantile function
\(f_{0}\) when \(\mathbb{P}(\epsilon\leq0\mid\mathbf{x})=\tau\)---in
particular under exogeneity---but in general differs from it under
endogeneity. The pinball loss \(\mathcal{P}_{\tau}(\widehat{f}_{h})\)
is thus an out-of-sample measure of predictive conditional-quantile
fit rather than of structural excess risk; since \(f_{0}\) is
unobserved and all methods are evaluated under the identical
criterion, relative comparisons across methods remain informative.

\begin{figure}[H]
\centering
\includegraphics[width=0.8\textwidth]{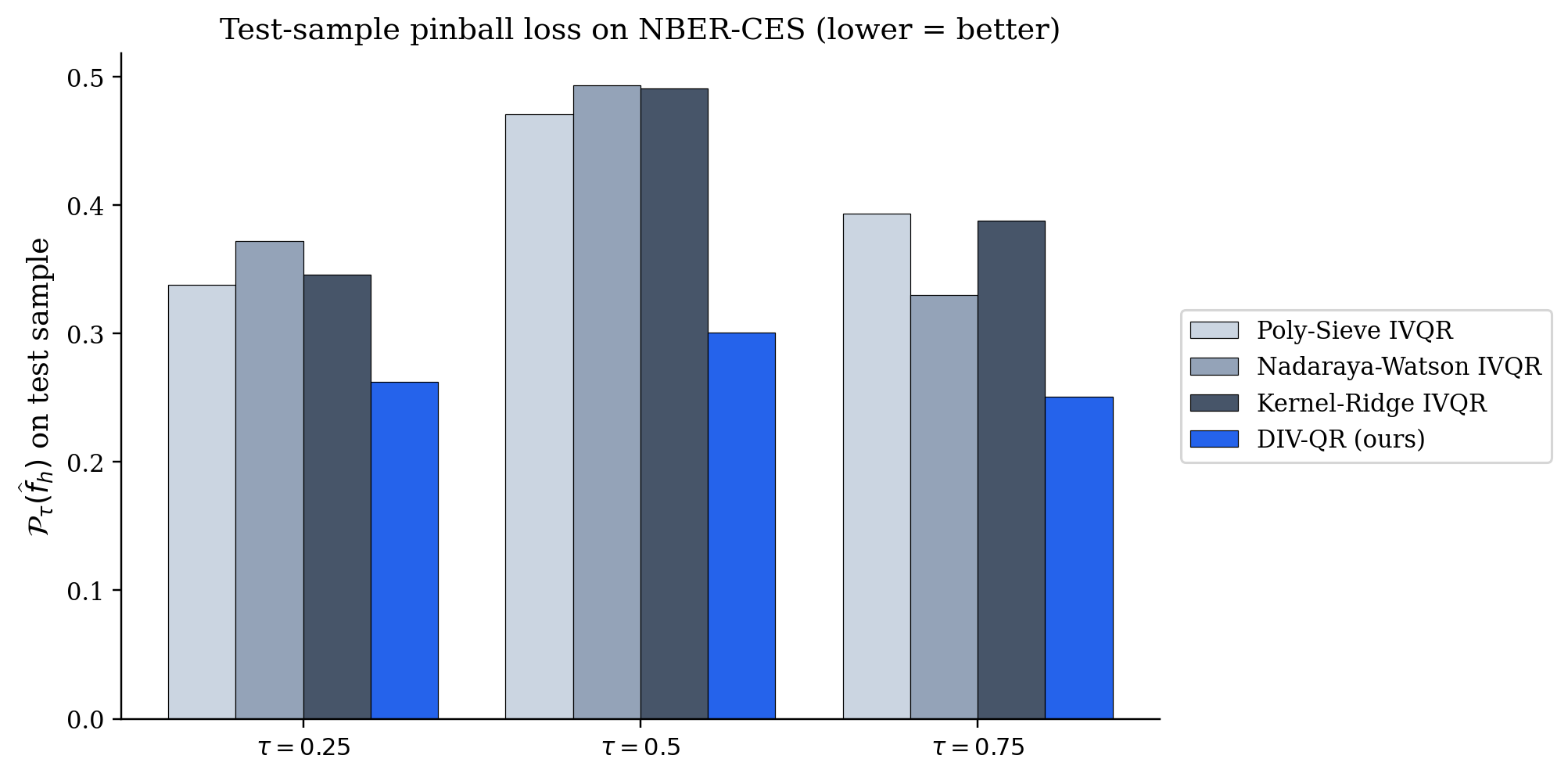}
\caption{Test-sample pinball loss
\(\mathcal{P}_{\tau}(\widehat{f}_{h})\) on the NBER-CES data. 
}
\label{fig:rda_pinball}
\end{figure}

\begin{figure}[H]
\centering
\includegraphics[width=0.9\textwidth]{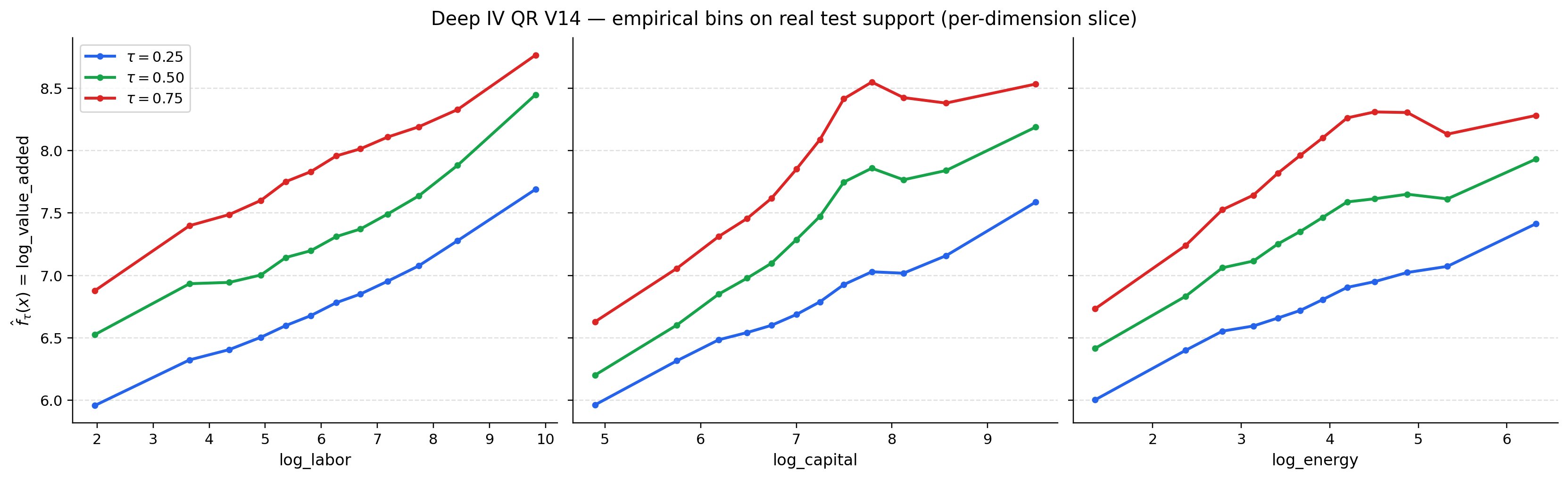}
\caption{Empirical slice curves of the fitted DIVQR surfaces
\(\widehat{f}_{h,\tau}(\mathbf{x})\) on the NBER-CES test sample.
}
\label{fig:rda_demand_curves}
\end{figure}

\noindent
\textbf
{Results.}
Table~\ref{tab:rda_main} and Figure~\ref{fig:rda_pinball} report the
test-sample pinball loss for each method and each target quantile.
At every \(\tau\in\{0.25,0.50,0.75\}\), DIVQR attains a strictly
lower pinball loss than each of the three baselines:
\(\mathcal{P}_{0.25}^{\,\mathrm{DIV\text{-}QR}}=0.262\) against a
best-baseline value of \(0.338\) (a reduction of \(22.5\%\));
\(\mathcal{P}_{0.50}^{\,\mathrm{DIV\text{-}QR}}=0.301\) against a
best-baseline value of \(0.471\) (a reduction of \(36.1\%\)); and
\(\mathcal{P}_{0.75}^{\,\mathrm{DIV\text{-}QR}}=0.251\) against a
best-baseline value of \(0.330\) (a reduction of \(24.0\%\)). Because
the pinball loss is an out-of-sample measure of conditional
\(\tau\)-quantile fit, these reductions translate into a
substantially better predictive fit of the fitted quantile surface
across the support of \(\mathbf{x}\). The four methods share the same stage-2
MLP function class, the same bandwidth,
the same loss function, and the same optimizer; the pinball advantage
of DIVQR therefore cannot originate in any of these components.
What remains to explain the gap is the one component that differs
across methods, namely the way the inner conditional expectation
\(\mathbb{E}[\,\cdot\,\mid\mathbf{z}]\) is estimated. The three
baselines each impose a particular nonparametric approximation, a
twenty-one dimensional polynomial basis, a Gaussian local smoother,
or a finite-dimensional RKHS model, whose approximation bias the
\(\widetilde{m}=1024\) diffusion samples employed by DIVQR do not share. The
magnitude and consistency of the pinball-loss gap, across three
different classical conditional-expectation estimators and three
different quantile levels, is indicative that access to samples from
the full conditional distribution 
\(\widehat{p}_{\mathbf{x},y\mid\mathbf{z}}\)
offers a tangible estimation advantage on this dataset.
Figure~\ref{fig:rda_demand_curves} provides a qualitative complement:
the three fitted DIVQR quantile surfaces are monotone in \(\log L\)
and \(\log K\), are clearly separated across \(\tau\), and accommodate
the negative partial dependence on \(\log E\) without imposing any
sign restriction.

\begin{table}[H]
\centering
\caption{Test-sample pinball loss
\(\mathcal{P}_{\tau}(\widehat{f}_{h})\) on the NBER-CES data.
}
\label{tab:rda_main}
\small
\renewcommand{\arraystretch}{1.15}
\begin{tabular*}{\textwidth}{@{\extracolsep{\fill}}l*{3}{c}}
\toprule
 & \(\mathcal{P}_{\tau}\) at \(\tau=0.25\)
 & \(\mathcal{P}_{\tau}\) at \(\tau=0.50\)
 & \(\mathcal{P}_{\tau}\) at \(\tau=0.75\)\\
\midrule
Poly-Sieve IVQR
 & \(0.338\) & \(0.471\) & \(0.393\)\\
Nadaraya-Watson IVQR
 & \(0.372\) & \(0.493\) & \(0.330\)\\
Kernel-Ridge IVQR
 & \(0.346\) & \(0.490\) & \(0.388\)\\
\textbf{DIVQR (ours)}
 & \(\mathbf{0.262}\) & \(\mathbf{0.301}\) & \(\mathbf{0.251}\)\\
\bottomrule
\end{tabular*}
\end{table}

\section{Conclusion}\label{sec:conclusion}

This work presents Deep nonparametric IVQR, a
two-stage nonparametric estimator for the structural quantile
function in IVQR models, built upon a variance-preserving
conditional diffusion model. By replacing the classical kernel or
sieve approximation of the conditional moment operator with a
learned conditional diffusion sampler and minimizing, over a deep
network class, the squared deviation between a kernel-smoothed
empirical moment and the target quantile level, DIVQR
remains effective in high-dimensional settings where existing
nonparametric IVQR methods break down. We provide rigorous
theoretical guarantees, including an end-to-end total variation
convergence rate for the conditional diffusion model under
unbounded support and non-asymptotic excess-risk bounds for the
resulting IVQR estimator, and we corroborate these guarantees by
simulation studies and a real-data analysis.
All guarantees are established under a polynomial-tail envelope
on the conditional density, and the theory degenerates continuously to
the exponential (Gaussian-envelope) regime: as the tail index grows,
the excess-risk rate converges to the minimax-optimal rate
$n^{-2\beta_q/(d+2\beta_q)}$ of nonparametric regression
and the first-stage rate recovers exactly the Gaussian-envelope
total-variation rate, so the proposed heavy-tailed theory covers the
classical light-tailed nonparametric guarantees as a limiting case.
{Future research may
extend the framework to inference on the structural quantile
function via cross-fitting and Neyman orthogonalization, to the
entire quantile process under 
non-crossing constraints, and to
adaptive procedures that do not require prior knowledge of the
smoothness index or the tail behaviour of the conditional density.
}


\bibliography{ref_bib}

\end{document}